\documentclass[10pt,twocolumn,letterpaper]{article}

\usepackage{iccv}              %

\definecolor{iccvblue}{rgb}{0.21,0.49,0.74}
\usepackage[pagebackref,breaklinks,colorlinks,allcolors=iccvblue]{hyperref}
\usepackage[accsupp]{axessibility}  %

\usepackage{booktabs}       %
\usepackage{multirow}
\usepackage{amsmath,amssymb}
\usepackage{color}
\usepackage{caption}
\usepackage{subcaption}
\usepackage{graphicx}
\usepackage{xspace}
\usepackage{bbm}
\usepackage{makecell}
\usepackage{nccmath}
\usepackage{xcolor}
\usepackage{csquotes}
\usepackage{pifont}%
\usepackage{fontawesome}
\usepackage{algorithm}
\usepackage{algorithmicx}  
\usepackage[noend]{algpseudocode}

\usepackage[capitalize]{cleveref}

\usepackage{tcolorbox}
\tcbuselibrary{breakable}

\newcommand{\piref}{\pi_\text{ref}}

\makeatletter
\DeclareRobustCommand\onedot{\futurelet\@let@token\@onedot}

\def\@onedot{\ifx\@let@token.\else.\null\fi\xspace}

\def\eg{\emph{e.g}\onedot} 
\def\ie{\emph{i.e}\onedot}

\def\etal{\emph{et al}\onedot}

\crefname{section}{Sec.}{Secs.}
\Crefname{section}{Section}{Sections}
\Crefname{table}{Table}{Tables}
\crefname{table}{Tab.}{Tabs.}
\crefname{figure}{Fig.}{Figs.}

\definecolor{myGreen}{rgb}{0,0.5,0.3}
\definecolor{myRed}{rgb}{0.6,0,0}

\def\paperID{12736} %
\def\confName{ICCV}
\def\confYear{2025}

\title{SCRAMBLe : Enhancing Multimodal LLM Compositionality with Synthetic Preference Data}

\author{
Samarth Mishra \qquad Kate Saenko \qquad Venkatesh Saligrama \\
Boston University \\
\small{\texttt{\{samarthm, saenko, srv\}@bu.edu}}
}

\begin{document}
\maketitle
\begin{abstract}
Compositionality, or correctly recognizing scenes as compositions of atomic visual concepts, remains difficult for multimodal large language models (MLLMs). Even state of the art MLLMs such as GPT-4o can make mistakes in distinguishing compositions like ``dog chasing cat'' vs ``cat chasing dog''. While on Winoground, a benchmark for measuring such reasoning, MLLMs have made significant progress, they are still far from a human's performance. We show that compositional reasoning in these models can be improved by elucidating such concepts via data, where a model is trained to prefer the correct caption for an image over a close but incorrect one. We introduce SCRAMBLe: Synthetic Compositional Reasoning Augmentation of MLLMs with Binary preference Learning, an approach for preference tuning open-weight MLLMs on synthetic preference data generated in a fully automated manner from existing image-caption data. SCRAMBLe holistically improves these MLLMs' compositional reasoning capabilities which we can see through significant improvements across multiple vision language compositionality benchmarks, as well as smaller but significant improvements on general question answering tasks. As a sneak peek, SCRAMBLe tuned Molmo-7B model improves on Winoground from 49.5\% to 54.8\% (best reported to date), while improving by ~1\% on more general visual question answering tasks. Code for SCRAMBLe along with tuned models and our synthetic training dataset is available at \href{https://github.com/samarth4149/SCRAMBLe}{https://github.com/samarth4149/SCRAMBLe}.

\end{abstract}

\section{Introduction} \label{sec:intro}

\begin{figure}[t!]
    \centering
    \includegraphics[width=\linewidth,trim=0cm 0cm 0cm 0cm,clip]{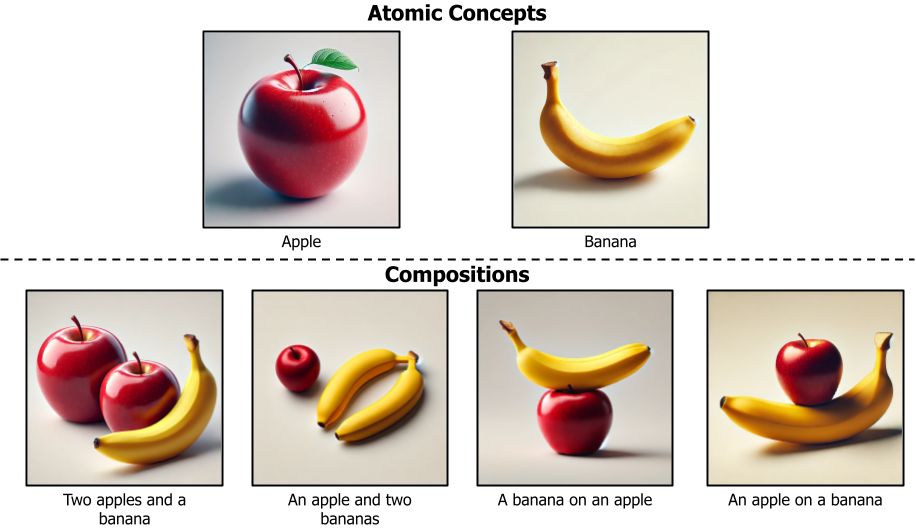}
    \vspace{-6mm}
    \caption{\textbf{Compositionality} or recognizing complex scenes as compositions of atomic concepts, is integral to human cognition~\cite{lake2017building}, and therefore desirable in vision language models.}
    \label{fig:cr_illustration}
    \vspace{-4mm}
\end{figure}

Compositionality in vision language models allows them to recognize a scene as a composition of its parts. With this if a model understands atomic concepts, it understands infinite compositions of them without additional representational capacity~\cite{lake2017building} (see \cref{fig:cr_illustration}).

\begin{figure}[h!]
    \centering
    \includegraphics[width=\linewidth,trim=0cm 0cm 0cm 0cm,clip]{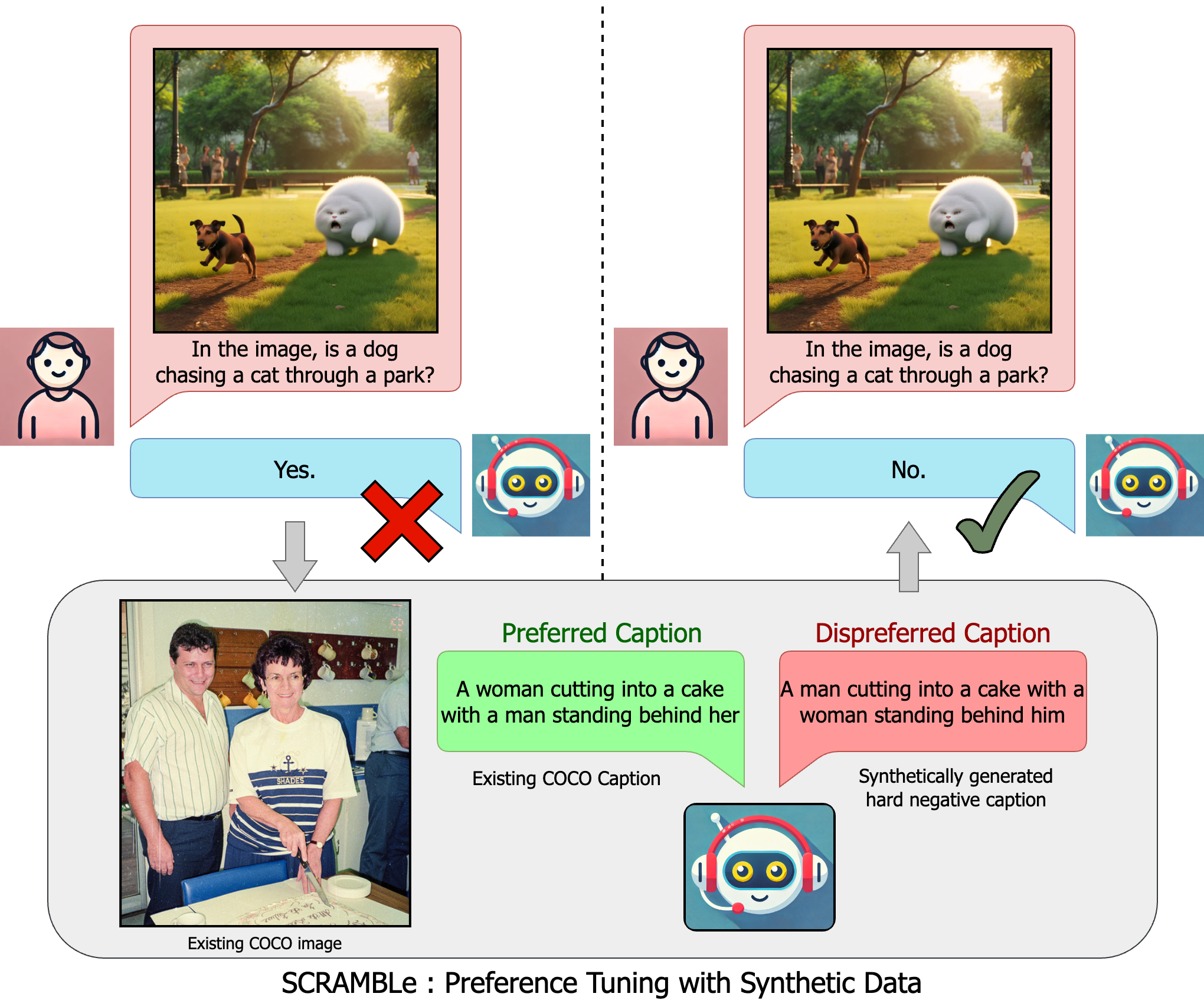}
    \caption{\textbf{SCRAMBLe.} Even state of the art multimodal large language models (MLLMs) (like GPT-4o) can falter on tests of compositionality, \eg determining the correct relation between the dog and the cat in the image. We propose SCRAMBLe, a fully automated approach of collecting synthetic preference data and tuning MLLMs for improving their significantly on compositionality, while also leading to smaller overall improvements as measure by general QA benchmarks.}
    \label{fig:scramble}
    \vspace{-8mm}
\end{figure}

A common test of compositionality assumes understanding of atomic concepts and assesses if a vision language model can determine how they are composed in a complex scene. Such a test reveals that even state of the art models struggle with compositionality, for \eg, GPT-4o~\cite{openai2024gpt4o} failed at correctly determining the relation between a dog and a cat in the image shown in \cref{fig:scramble}. 
To tackle such failures, we introduce SCRAMBLe, a novel approach that automates high-quality synthetic negative caption generation for preference tuning MLLMs. While prior work synthetically generating hard negatives often produce grammatically or logically incorrect captions, a trait models can use to learn shortcuts, we improve quality of hard negatives generated using LLM reasoning and subsequent plausibility and grammar based filtering, ensuring that negatives remain syntactically and semantically valid while still presenting a meaningful challenge for training.

\noindent \textbf{Evaluating Compositionality.} Winoground~\cite{thrush2022winoground} is a seminal benchmark that uncovered such shortcomings in compositionality of large-scale pre-trained vision-language models. A model is given two images and two captions and is asked to match each image to its caption and vice versa. This evaluates compositionality in the manner that the two captions use the same set of words but paint very different pictures, often by changing the relation between objects, but sometimes by changing meanings of words. Besides Winoground, other benchmarks (\eg COLA~\cite{ray2024cola}, EqBen~\cite{wang2023equivariant}, ConME~\cite{huang2024conme}) have been used as tests of compositional reasoning focusing on recognition when atomic concepts, their properties or their relations change in complex scenes. Refer to \cref{fig:benchmark_examples,,fig:conme_examples} for examples from these benchmarks.

\noindent \textbf{Performance of Vision Language Models.} Starting with models such as CLIP~\cite{radford2021learning}, that were shown to have atrociously low performance (\eg below chance on Winoground~\cite{thrush2022winoground}), vision-language models have improved significantly by using large language model foundations (as multimodal large language models or MLLMs). For instance, LLaVA~\cite{liu2024improved} has been reported to have a Winoground accuracy (group score) of 30\%~\cite{lin2024evaluating} up from CLIP's 8\%. There is however lots of room for improvement when it comes to compositionality of these models.

\noindent \textbf{Improving Compositionality.} Some works have found that better compositionality can be realized by spending more inference time compute with these MLLMs to reason better through the answering process. For instance, Mitra~\etal~\cite{mitra2024compositional} and Zhang~\etal~\cite{zhang2024cocot} have found different chain of thought prompting methods can help MLLMs improve performance on compositionality benchmarks. Along similar lines, CECE~\cite{cascante2025natural} is a method that with the help of an LLM, expands a caption into a larger set of entailments and contradictions, and performs the matching task based on this expanded set. While these methods are effective, a good MLLM should not have to spend time generating such entailments to answer simple questions about objects comprising an image, their properties and/or relations between them. 

\noindent \textbf{SCRAMBLe.} With this motivation, we propose to improve MLLMs via preference tuning, to teach them to disprefer failure modes such as complex scene captions that are mostly correct but incorrect in a small/subtle way. While it may be possible to collect many more high quality captions of complex scenes with expensive labeling pipelines to improve MLLM compositionality, we chose this approach since our data curation pipeline is fully automated (hence comparatively inexpensive) using existing image-caption data that MLLMs may have already been trained on (\eg COCO~\cite{lin2014microsoft}).

Conceptually our approach is similar to NegCLIP~\cite{yuksekgonul2022and} where hard negative captions are generated by programmatically swapping nouns/adjectives/verbs/adverb phrases in the original caption for tuning a CLIP model. This approach has also been used for creating evaluation benchmarks~\cite{zhao2022explainable, ma2023crepe}, but it was Hsieh~\etal~\cite{hsieh2024sugarcrepe} who first pointed that if negative captions are non-grammatical or illogical, models can use shortcuts to beat them. We argue that when such data is used for training, it may teach a model only trivial differences between preferred and dispreferred outcomes. Hence, as mentioned above, we focus on high quality hard negative caption generation by using an LLM expert's reasoning, and filtering to discard low quality (\eg non-grammatical or illogical) captions. 

This simple approach substantially improves MLLM performance, not just for compositionality, but holistically. As a sneak peek, on a Molmo-7B~\cite{deitke2024molmo} model, our SCRAMBLe tuning approach leads to a performance of 54.8\% (up from 49.5\%), the best reported on Winoground to date. This is achieved along with ~1\% improvement on general question answering benchmarks such as SEED-Bench~\cite{li2023seed} and MM-VET~\cite{yu2023mm}. We demonstrate SCRAMBLe's efficacy using 6 compositionality and general QA benchmarks with 3 recent open-weight MLLMs. We also show SCRAMBLe's data quality is higher and leads to better performance than prior hard negative caption generation methods.

\section{Background and Related Work} \label{sec:related}

\noindent \textbf{Vision Language Models.} Large-scale pre-training unlocked many-a-capability in different vision language models~\cite{bommasani2021opportunities}. CLIP~\cite{radford2021learning} is one such model with separate image and text encoders trained to match features of images to features of their textual descriptions. This model can then be used for a range of tasks such as zero-shot classification~\cite{radford2021learning,pratt2023does,zhou2022learning}, text-to-image retrieval~\cite{saito2023pic2word}, image or text similarity measurement~\cite{hessel2021clipscore}, text to image generation~\cite{ramesh2022hierarchical,rombach2022high} among others. CLIP also found another very important application in combination with large language models~\cite{touvron2023llama,chiang2023vicuna} to create a general purpose image question answering chat agent~\cite{zhu2023minigpt,liu2024visual,liu2024improved}. More capable open weight MLLMs have subsequently followed similar core architectures~\cite{agrawal2024pixtral,deitke2024molmo,dubey2024llama} and closed the gap to proprietary models like GPT-4~\cite{achiam2023gpt}. As the more capable class of models, we are interested in their compositionality, which we show can be significantly improved with our approach.

\noindent \textbf{Visio-Linguistic Compositionality.} Compositionality is central to human intelligence and allows us to comprehend infinite compositions of finite numbers of concepts~\cite{lake2017building}. It is hence a desirable in vision language models too. Thrush~\etal~\cite{thrush2022winoground} conducted a controlled experiment to test this and found
CLIP has worse than chance performance on their dataset, alongside other image and text representation learning approaches~\cite{chen2020uniter,zhang2021vinvl}. Subsequently Mitra~\etal~\cite{mitra2024compositional} and Lin~\etal~\cite{lin2024evaluating} proposed ways of evaluating MLLMs on this benchmark (with and without access to model output probability distributions repectively) and found them to be much better than models like CLIP. Diwan~\etal~\cite{diwan2022winoground} in their study of Winoground, found some examples may be hard for reasons other than lack of compositional reasoning capability (\eg ambiguity, visual difficulty of identification, etc.). Besides Winoground, we evaluated using other benchmarks like EqBen~\cite{wang2023equivariant}, Cola~\cite{ray2024cola} and ConME~\cite{huang2024conme} for additional compositionality testing scenarios.

\noindent \textbf{Improving Compositionality.} After Thrush~\etal's~\cite{thrush2022winoground} study, multiple benchmarks used synthetically generated hard negative captions for evaluating compositionality~\cite{yuksekgonul2022and, zhao2022explainable, ma2023crepe}.
Hsieh~\etal pointed out~\cite{hsieh2024sugarcrepe} that if the quality of these hard negatives is poor grammatically or logically, text-only models could exploit this bias to pick the correct caption over the negative one. From these works, Yuksekgonul~\etal~\cite{yuksekgonul2022and} also proposed NegCLIP, \ie tuning CLIP with their generated hard negatives, but this did not lead to much better performance on Winoground (as reported in \cite{herzig2023incorporating}). Our approach, SCRAMBLe, is similar conceptually to NegCLIP, except with an application to MLLMs and an emphasis on the quality of generated hard negative captions. We compare with a baseline data generation approach based on SugarCREPE~\cite{hsieh2024sugarcrepe}, a quality-focused evaluation benchmark based on synthetic hard negatives, and find that our generation leads to higher quality hard negatives, and hence to better performing tuned models.

Besides the above, there have been a range of other approaches attempting to improve CLIP's compositionality : Tuning with additional synthetic images and captions~\cite{cascante2023going}, with losses to improve equivariance~\cite{herzig2023incorporating}, with dense and aligned captions~\cite{doveh2023dense}, or with structured captions rewritten by LLMs~\cite{doveh2023teaching}. These approaches can be translated to MLLMs, and may be complementary to our approach. An investigation of them is left for future work.

As mentioned in the introduction, in MLLMs, prior compositionality improvement have been made with additional test time compute to elicit more/better reasoning~\cite{mitra2024compositional,zhang2024cocot,cascante2025natural}. SCRAMBLe, in contrast, attempts to improve MLLMs directly by teaching them distinctions with high quality preference data. In this manner, it is complementary to the aforementioned approaches.

\noindent \textbf{Preference Optimization of Language Models.} Human preferences over pairs of outcomes are a powerful and relatively easy to collect signal, which has been shown effective in training models for following instructions~\cite{ouyang2022training}, summarization~\cite{stiennon2020learning}, translation~\cite{kreutzer2018reliability} and story-telling~\cite{ziegler2019fine}. These works trained a separate model to learn a reward function compatible with human preferences, that was then used to fine-tune the language model with reinforcement learning. Rafailov~\etal~\cite{rafailov2024direct} proposed a training procedure that bypasses the reward model and fine-tunes a language model directly on preference data. As a stable and computationally efficient training procedure, we use this method for fine-tuning MLLMs with our synthetic preference data.

\section{Method} \label{sec:method}
Our approach, SCRAMBLe, introduces a fully automated pipeline for generating high-quality synthetic preference data to improve MLLMs’ compositional reasoning. Unlike prior approaches that swap phrases without always checking for coherence (\eg NegCLIP), SCRAMBLe incorporates grammar and plausibility filtering to ensure the generated negatives are both challenging and logically sound.
 
\subsection{Synthetic Data Generation} \label{sec:synth-gen}

\noindent\textbf{Baseline : Swap Objects/Attributes.}
Here we describe a baseline data generation approach similar to SugarCREPE~\cite{hsieh2024sugarcrepe} swap. Here, hard negatives are generated by swapping nouns or adjectives in the input caption, when possible. The specific prompt used for swapping nouns is shown in \cref{fig:swap-obj-prompt}. While an effective method, we found that swaps generated would still often be nonsensical. There were also many cases when two nouns were swapped around phrases like ``and'', ``next to'' or ``beside'' and the resulting negative caption would have the same meaning as the original caption.

\begin{figure}
    \centering
    \fcolorbox{black}{gray!10}{%
    \parbox{0.95\linewidth}{%
        \ttfamily
        \fontsize{8}{11}\selectfont
        Given an input sentence describing a scene, your task is to first locate two swappable noun phrases in the sentence, and then swap them to make a new sentence. The new sentence must meet the following three requirements:

        \begin{enumerate}
            \item The new sentence must be describing a different scene from the input sentence.
            \item The new sentence must be fluent and grammatically correct.
            \item The new sentence must make logical sense.
        \end{enumerate}
        
        To complete the task, you should:
        \begin{enumerate}
            \item Answer the question of whether generating such a new sentence is possible using Yes or No.
            \item Output the swappable noun phrases.
            \item Swap the selected noun phrases to generate a new sentence.
        \end{enumerate}

        Input: A woman cutting into a cake with a man standing behind her.

        Is it possible to swap noun phrases in the input sentence to generate a new sentence that is different from the input sentence and makes logical sense? \textcolor[rgb]{0, 0.5, 0}{Yes.\\
        Swappable noun phrases: a woman, a man\\
        Output: A man cutting into a cake with a woman standing behind him.
        }
        }%
    }
    \caption{\textbf{Baseline Hard-negative Generation : Swap Objects.} Example prompt to swap noun phrases and generated output (green) from the Llama-3.1 expert.}
    \label{fig:swap-obj-prompt}
    \vspace{-3mm}
\end{figure}

\noindent\textbf{Our improved approach with Chain of Thought.} In constrast to the baseline, we get the LLM to reason more generally through the negative caption generation process, by providing a chain of thought in the output~\cite{wei2022chain}. The initial prompt to the LLM mentions the same requirements as \cref{fig:swap-obj-prompt}, with the distinction that we now ask the LLM to rearrange words as it sees fit. To help with the reasoning process, we provide the LLM expert with in-context examples. One such example (out of the 5 we provide) is in \cref{fig:cot-prompt}. We source these examples also automatically by choosing 5 caption pairs (no images) from Winoground and asking Claude 3.5 Sonnet~\cite{anthropic2024introducing} how it would reason through if it was given the task of transforming one caption to its counterpart under the 3 requirements in \cref{fig:swap-obj-prompt}.

\begin{figure}
    \centering
    \fcolorbox{black}{gray!10}{%
    \parbox{0.95\linewidth}{%
        \ttfamily
        \fontsize{8}{11}\selectfont

        Input Caption: a pink bird with a white beak \\
        Reasoning: 
        \begin{enumerate}
            \item Identify the key elements: 
            \begin{itemize}
                \item Color of the bird: pink
                \item Color of the beak: white
                \item Structure: "[color] bird with a [color] beak"
            \end{itemize}
            \item Recognize that the negative caption should describe a different visual image using the same words
            \item Observe that the colors are the main distinguishing features
            \item Swap the colors while maintaining the structure:
            \begin{itemize}
                \item "pink" describing the bird becomes "white"
                \item "white" describing the beak becomes "pink"
            \end{itemize}
            \item Keep the rest of the structure intact:
            \begin{itemize}
                \item "a [color] bird with a [color] beak"
            \end{itemize}
            \item Apply the swapped colors to the structure:
            \begin{itemize}
                \item "a white bird with a pink beak"
            \end{itemize}
        \end{enumerate}
        Final Output Caption : a white bird with a pink beak

        }%
    }
    \caption{A \textbf{Reasoning Chain of Thought} for transforming a caption into a negative caption. We provide 5 such examples in-context to the Llama-3.1 LLM expert to help it reason through the process of generating a new negative caption.}
    \label{fig:cot-prompt}
    \vspace{-4mm}
\end{figure}

\subsection{Filtering} \label{sec:filtering}
We note the negative caption generation method we described is noisy and susceptible to making errors. Hence, we use a post filtering step to keep only a subset of high quality negative captions. We use the Vera plausibility model~\cite{liu2023vera} and the grammar model from TextAttack~\cite{morris2020textattack} in an adversarial refinement process (similar to \cite{hsieh2024sugarcrepe}) for this step. This process is aimed at debiasing the generated dataset based on the score differences between positive and negative captions. It ensures there are an equal number of examples with a certain score difference as there are with the negative of that score difference, \ie, the distribution of score differences is symmetric about zero. This debiasing prevents a model trained on this data from learning shortcuts using the plausibility or grammatical correctness of the captions (and disregarding images). For the sake of brevity, the adversarial refinement procedure is described in full detail in \cref{sec:app-adv_ref}.

We also note that while adversarial refinement debiases the dataset, it also acts as a filter of low quality captions in terms of grammar and plausibility. Most positive captions have fairly high plausibility and grammar scores, given the images are captioned by humans. Hence, if negative captions with low scores were to be retained, to keep the dataset unbiased, an equal number of examples would be needed with the negative score difference, \ie, examples where the negative caption is a lot more plausible and grammatically correct compared to the positive caption. Such examples are extremely rare and in some cases impossible because the upper bound on the scores is 1.

\subsection{Training} \label{sec:training}
Once we have a preference dataset ($\mathcal{D}$) of images, their captions and hard negative captions, we fine-tune a low-rank adapter~\cite{hu2021lora} on a target MLLM with the direct preference optimization~\cite{rafailov2024direct} objective: 
\begin{align}\label{eq:dpo_objective}
    \mathcal{L}_\text{DPO}(\theta) = -\mathbb{E}_{(x, y_w, y_l)\sim \mathcal{D}}&\left[\log \sigma \left( \beta \log \frac{\pi_{\theta}(y_w\mid x)}{\piref(y_w\mid x)}\right.\right. \nonumber \\ 
    &\left.\left. -  \beta \log \frac{\pi_{\theta}(y_l\mid x)}{\piref(y_l\mid x)}\right)\right].
\end{align}
where $x$ is a prompt, in our case, an image and the prompt ``caption : '', $y_w$ is the preferred outcome \ie the image caption, and $y_l$ is the dispreferred hard negative caption. We use the MLLM before finetuning as the reference model $\piref$. $\pi_\theta$ is the model with the adapter ($\theta$ being the trainable adapter parameters). $\beta$ is a parameter controlling the deviation from the reference model and $\sigma$ the sigmoid function. Note that $\pi(y\mid x)$ is the probability of generating the caption $y$ given the prompt $x$ from the model $\pi$.

\section{Experiments}

\subsection{Evaluation.} \label{subsec:evaluation}

\begin{figure*}[t]
    \centering
    \includegraphics[width=0.95\textwidth]{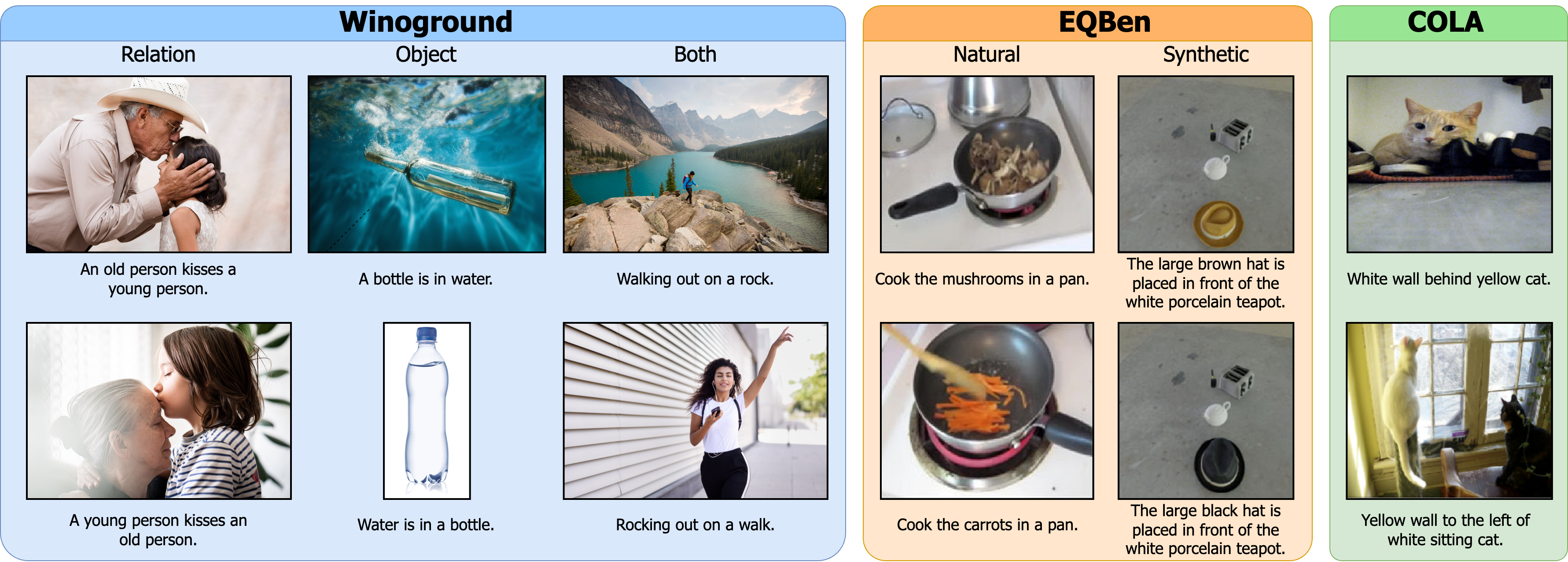}
    \caption{\textbf{Examples from compositionality benchmarks.} Winoground has examples in natural images swapping objects, relations or both. EqBen has both natural and synthetic images in examples where typically a small aspect changes between two images. Cola has examples from natural images with different relations between two objects.}
    \label{fig:benchmark_examples}
    \vspace{-4mm}
\end{figure*}

\begin{figure}[h!]
    \centering
    \includegraphics[width=0.8\linewidth,trim=0cm 0cm 0cm 0cm,clip]{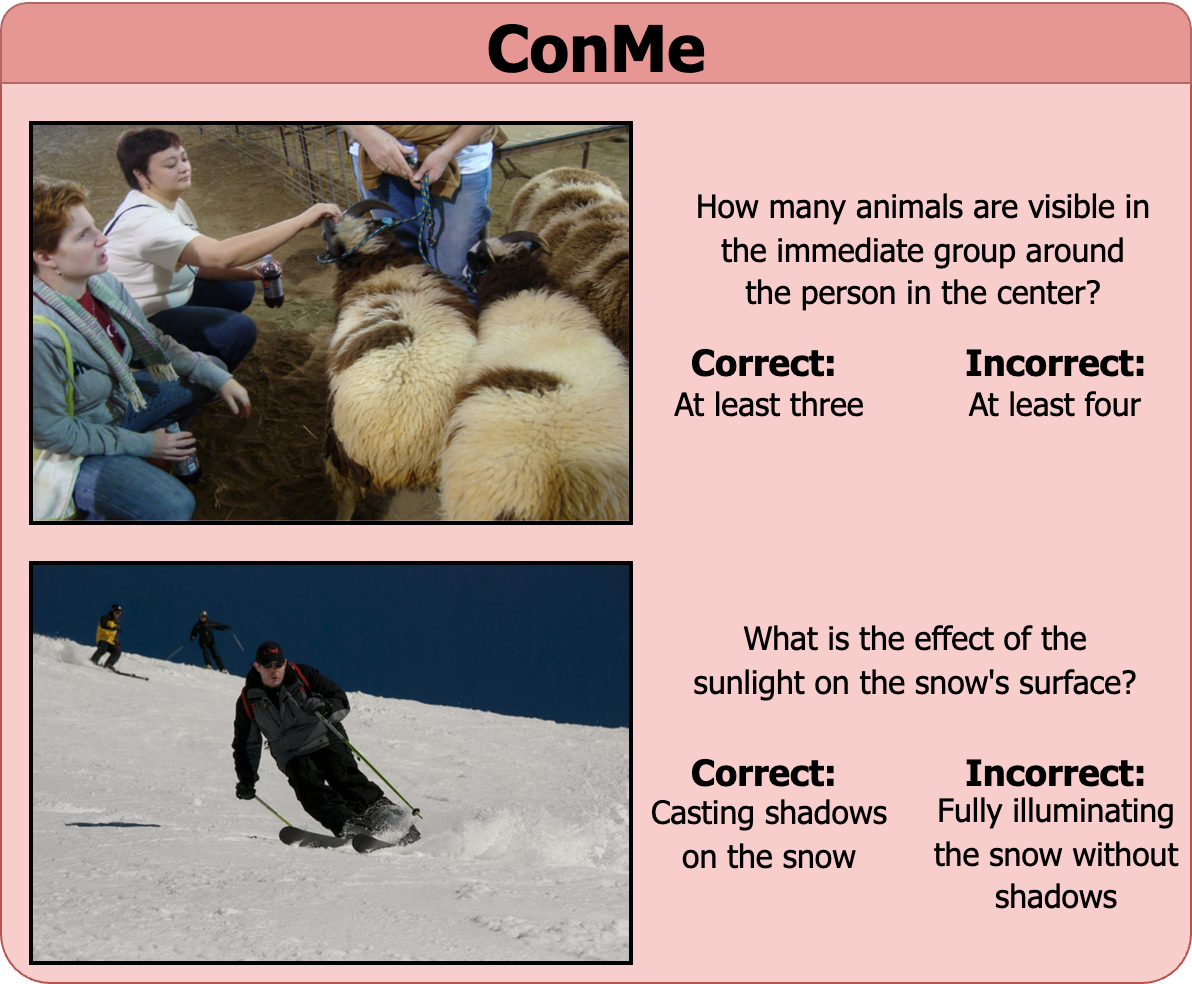}
    \vspace{-2mm}
    \caption{\textbf{Examples from ConMe.} ConMe is another compositionality test benchmark that poses a two choice question to an MLLM, asking it to output A or B for it's choice.}
    \label{fig:conme_examples}
    \vspace{-5mm}
\end{figure}

We evaluated the performance of the tuned MLLMs on four compositionality benchmarks (Winoground~\cite{thrush2022winoground}, EqBen~\cite{wang2023equivariant}, Cola~\cite{ray2024cola} and ConMe~\cite{huang2024conme}) and two other image question answering benchmarks (SEED-Bench~\cite{li2023seed} and MM-Vet~\cite{yu2023mm}), used as control to ensure the tuned MLLMs do not lose performance on more general tasks. 

\noindent\textbf{Compositionality Benchmarks.} An example from ConMe consists of 1 image based on which the MLLM is asked a question with 2 options denoted with ``A'' or ``B'' , and it is expected to generate the letter corresponding to the correct option (see examples in \cref{fig:conme_examples}). 

The remaining three benchmarks involve a matching task with 2 images and 2 captions in each example (see examples in \cref{fig:benchmark_examples}). Winoground examples could involve object swaps, relation swaps or both. While the first reorders elements such as noun phrases referring to objects in the image, relation swaps reorder elements such as verbs, adjectives, prepositions, and/or adverbs. Swaps of both objects and relations could lead to meanings of words changing across the captions. EqBen is a dataset consisting of both real and synthetic images, with typically a small aspect changed between two images. Cola has examples from natural images with different relations between two objects. An example is deemed correctly answered if for each image, the model prefers its caption over the other and vice versa for each caption (this corresponds to the group score accuracy defined in \cite{thrush2022winoground}).

\noindent\textbf{Image-Caption Affinity metric (VQAScore).} In the compositionality benchmarks with 2 images and captions each, a model can be scored if it can provide a preference/affinity metric between an image and a caption. For this, we use VQAScore~\cite{lin2024evaluating}, where a model's score for an image-caption pair is the probability, $\pi_{\theta}(y \mid x)$, of generating the output $y=$ ``Yes'' when asked the question $x=$ ``Does this image show $\langle$caption$\rangle$?''. 
Note that this requires the output probability distribution over a model's vocabulary tokens which is not typically available for closed-source models. 

\noindent\textbf{Control Benchmarks.} SEED-Bench~\cite{li2023seed} is a large evaluation set of multiple choice questions over both images and videos, of which we only use the image subset containing 14.2k questions. It contains questions testing a variety of capabilities such as scene understanding, instance identity/attribute identification, counting and visual reasoning among others. MM-Vet~\cite{yu2023mm} is a dataset of 218 questions where the model has to generate a descriptive answer. The generated answer is compared with a ground truth answer by a GPT-4 judge and scored between 0 and 1. The questions test abilities like recognition, optical character recognition, general knowledge, spatial reasoning and math.

\begin{table*}[t]
    \centering
    \scalebox{0.9}{
    \begin{tabular}{@{}l|cccc|cc@{}}
        \toprule
        & \multicolumn{4}{c|}{\textbf{Compositionality Benchmarks}} & \multicolumn{2}{c}{\textbf{Control Benchmarks}} \\
        \midrule
        \textbf{Model   Name} & \textbf{Winoground} & \textbf{EqBen} & \textbf{COLA} & \textbf{ConMe} & \textbf{SEED-Bench} & \textbf{MM-Vet} \\ \midrule
        LLaVA-1.5-13B & 36.5 & 36.4 & 49.5 & 62.3 & \textbf{68.23} & 36.2 $\pm$ 0.3 \\
        +   SCRAMBLe & \textbf{39.3} & \textbf{39.3} & \textbf{55.7} & \textbf{64.5} & 68.19 & \textbf{38.6 $\pm$ 0.1} \\
        MoLMo-7B & 49.5 & \textbf{62.9} & 57.1 & 72.2 & 74.04 & 59.3 $\pm$ 0.2 \\
        +   SCRAMBLe & \textbf{54.8} & 59.3 & \textbf{60.5} & \textbf{74.6} & \textbf{74.61} & \textbf{60.9 $\pm$ 0.4} \\
        Llama-3.2-11B & 31.5 & 43.6 & 37.1 & 71.3 & 13.79 & 57.0 $\pm$ 0.1 \\
        +   SCRAMBLe$^{\ast}$ & \textbf{35.3} & \textbf{44.3} & \textbf{40.0} & \textbf{74.6} & \textbf{42.74} & \textbf{60.3 $\pm$ 0.1}\\ \bottomrule
    \end{tabular}
    }
    \caption{\textbf{SCRAMBLe improves performance of different open-weight MLLMs across compositionality and control benchmarks.} We note that LLama-3.2 shows significantly weaker performance on SEED-Bench, largely due to formatting issues rather than an inherent lack of capability. Refer to \cref{sec:results} for full discussion.}
    \label{tab:diff_models}
    \vspace{-2mm}
\end{table*}

\begin{table*}[h]
    \centering
    \scalebox{0.7}{
    \begin{tabular}{@{}p{8cm}|p{8cm}|p{8cm}@{}}
    \toprule
    \multicolumn{1}{c|}{\multirow{2}{*}{\textbf{Positive Caption}}} & \multicolumn{2}{c}{\textbf{Generated Negative Caption}} \\ \cmidrule{2-3}
    & \multicolumn{1}{c|}{\textbf{(Baseline) Swap Obj/Att}} & \multicolumn{1}{c}{\textbf{SCRAMBLe : Chain of Thought}} \\
    \midrule\midrule
    A red fire hydrant, a yellow balloon, and   some rocks. & (Att) A yellow fire hydrant, a red balloon, and some rocks & A yellow fire hydrant, a red balloon, and some rocks \\
    A woman cutting into a cake with a man   standing behind her & (Obj) A man cutting into a cake with a woman standing behind him & A man cutting into a cake with a woman standing behind him. \\
    A young child and two adults go skiing. & (Obj) A young adult and two children go skiing. & Two children and a young adult go skiing \\
    A young zebra is sniffing the ground in a   dusty area. & (Obj) \textcolor{red}{A young area is sniffing the ground in a dusty zebra.} & An old zebra is sniffing the ground in a dusty area. \\
    Two giraffes standing in a tree-filled   area. & (Att) \textcolor{red}{NA} & Two giraffes standing outside a tree-filled area. \\ \bottomrule
    \bottomrule
    \end{tabular}
    }
    \caption{\textbf{Qualitative examples} of negative captions generated by baseline vs our approach (red highlights abstentions/poor generations). Swap works well in certain cases when swaps are possible (first three). Chain of thought can handle cases outside of this with higher quality hard negatives.}
    \label{tab:qual_eg}
\end{table*}

\subsection{Implementation Details}
We ran all experiments on single 48G Nvidia Ampere gpus. For the LLM expert generating synthetic captions, we used a Llama-3.1-70B model with inference run at 4-bit (nf4) quantization\footnote{we found little performance difference between the quantized and non-quantized models}. 
We tuned and evaluated LLaVA-v1.5-13B~\cite{liu2024improved}\footnote{We experimented with LLaVA-v1.6 with both vicuna-13B and mistral-7B versions but found them to have poorer compositionality than LLaVA-v1.5}, Molmo-7B-D-0924~\cite{deitke2024molmo} and Llama-3.2-11B-Vision-Instruct~\cite{dubey2024llama} (where 13B, 7B and 11B refer to the approx. number of parameters of the models). 
We added a low-rank adapter (LoRA~\cite{hu2021lora}) to each model and trained at half (bfloat16) precision for 2 epochs with an effective batch size of 8 (using gradient accumulation). For full training details please refer to \cref{sec:app-impl}.

\begin{table}[h]
    \centering
    \scalebox{0.85}{
    \begin{tabular}{@{}l|cc|c@{}}
        \toprule
        \multicolumn{1}{l|}{\textbf{Stage}} & \multicolumn{2}{c|}{\textbf{(Baseline) Swap}} & \textbf{Chain of} \\
        \multicolumn{1}{c|}{} & \textbf{Obj} & \textbf{Att} & \textbf{Thought} \\ \midrule
        Start & \multicolumn{3}{c}{118287} \\ \midrule
        Initial   Generation & 110855 & 21813  & 107679 \\
        Post-Filtering & 15546 & 7904  & 57786 \\
        \midrule
        Final & \multicolumn{2}{c|}{23450} & 57786 \\
        \bottomrule
    \end{tabular}
    }
    \caption{\textbf{Number of examples} initially generated and those retained after adversarial refinement for each negative caption generation method.}
    \label{tab:num_examples}
    \vspace{-2mm}
\end{table}

\begin{table*}[t]
    \centering
    \scalebox{0.9}{
    \begin{tabular}{@{}ll|cccc|cc@{}}
    \toprule
    & & \multicolumn{4}{c|}{\textbf{Compositionality Benchmarks}} & \multicolumn{2}{c}{\textbf{Control Benchmarks}} \\ \midrule
    \multicolumn{1}{c}{\textbf{Model   Name}} & \multicolumn{1}{c|}{\textbf{Tuning Data}} & \multicolumn{1}{c}{\textbf{Winoground}} & \multicolumn{1}{c}{\textbf{EqBen}} & \multicolumn{1}{c}{\textbf{COLA}} & \multicolumn{1}{c|}{\textbf{ConMe}} & \textbf{SEED-Bench} & \multicolumn{1}{c}{\textbf{MM-Vet}} \\ \midrule
    LLaVA-1.5-13B & \multicolumn{1}{c|}{-} & 36.5 & 36.4 & 49.5 & 62.3 & 68.23 & 36.2 $\pm$ 0.3 \\
    Baseline (w tuning) & Swap Obj/Att & 38.8 & 36.4 & 52.9 & 64.4 & \textbf{68.49} & 30.7 $\pm$ 0.4 \\
    SCRAMBLe (Ours) & Chain of Thought & \textbf{39.3} & \textbf{39.3} & \textbf{55.7} & \textbf{64.5} & 68.19 & \textbf{38.6 $\pm$ 0.1} \\
    \bottomrule
    \end{tabular}
    }
    \caption{\textbf{SCRAMBLe vs baseline caption generation.} We find the chain of thought method to lead to better quality hard negatives and hence better final performance.}
    \label{tab:diff_gen}
    \vspace{-2mm}
\end{table*}

\subsection{Results} \label{sec:results}

\noindent\textbf{SCRAMBLe improves performance of different MLLMs over compositionality and control benchmarks.} We experimented with SCRAMBLe tuning three recent open-weight MLLMs and the results are reported in \cref{tab:diff_models}. For Winoground, EQBen and Cola, we report only the group score accuracy and for ConMe and SEED-Bench, we report average accuracy overall. For MM-Vet performance we report the average and 95\% confidence interval over 5 evaluations with a GPT-4 judge~\cite{yu2023mm}.

\begin{figure}[]
    \centering
    \includegraphics[width=\linewidth,trim=0cm 0cm 0cm 0cm,clip]{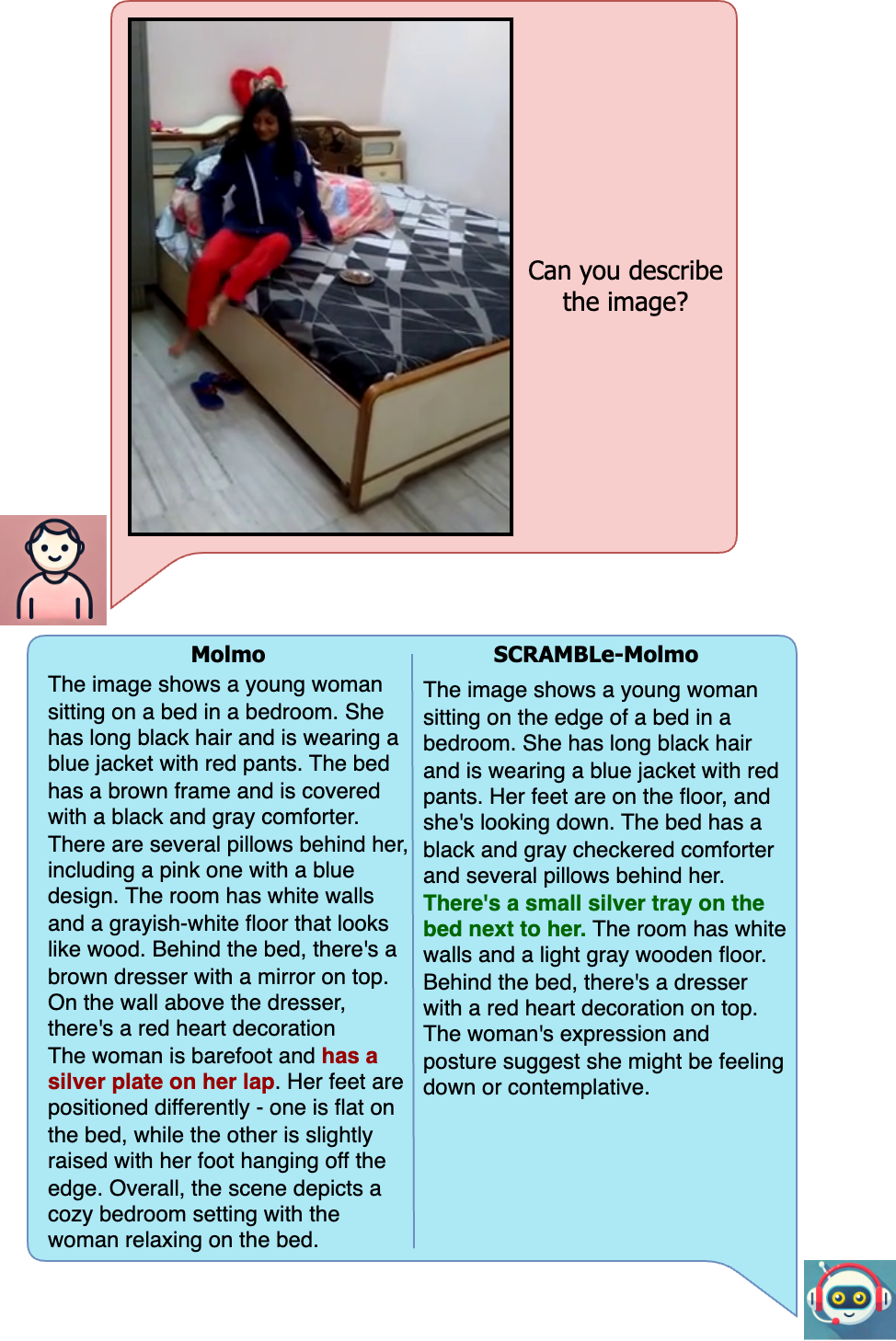}
    \caption{\textbf{Conversing with SCRAMBLe-Molmo (EQBen example).} SCRAMBLe can fix mistakes Molmo makes in identifying the location of the silver plate.}
    \label{fig:chat_eg1}
    \vspace{-2mm}
\end{figure}

All models improve in performance across most compositionality and general QA control benchmarks, showing that SCRAMBLe tuning is effective across MLLMs and can holistically improve their performance while focusing on compositionality. We note that while Molmo seems to become worse on EQBen, on qualitative evaluation on examples where Molmo and SCRAMBLe-Molmo disagree, we find that the former actually makes errors in recognition that the latter does not make (\eg in \cref{fig:chat_eg1}). On examples where SCRAMBLe-Molmo performed worse, we found no difference in recognition capabilities on interacting via chat (examples in \cref{sec:app-chat_examples}). We also note Llama-3.2 model's performance on SEED-Bench is quite low primarily because it often tended to explain solutions and did not follow the format instructions of answering precisely for the benchmark. While this highlights a format misalignment rather than a compositionality failure, it leaves open the question of whether better prompting strategies could improve performance without additional fine-tuning. As a second note, while tuning Llama-3.2, we found that it overfits when trained with the full dataset of 57.8k synthetic examples, but does improve when tuned on a smaller set ($\sim$10k examples). More discussion/results on this in \cref{sec:app-training}. 

\noindent\textbf{Comparing Caption Generation Procedures.} We show our chain of thought based hard negative generation leads to higher quality synthetic data than the baseline swap objects/attributes approach. Note that each method first involves prompting a Llama-3.1-70B LLM expert to get an initial set of captions, followed by adversarial filtering (\cref{sec:filtering}). In the first stage, the LLM is given an option to abstain from generation if it considers the task to be impossible. 

The number of examples we end up with from each generation method at the end of these two stages is shown in \cref{tab:num_examples}. While looking at the two generation methods of swapping objects and attributes, we notice that the LLM expert attempts to swap objects much more often than attributes. This however leads to a high number of low quality captions that get filtered out. We found that we could improve the initial generation quality a bit by using in-context examples (as done in \cite{hsieh2024sugarcrepe} and as we do for swapping attributes) but this typically resulted in many more abstentions, even when a swap of noun phrases was possible. We find that using reasoning with chain of thought, initial generations have a higher quality and a lot fewer of them are discarded in the filtering stage. 

Some examples of generated captions are shown in \cref{tab:qual_eg}. In the first three, we see that attributes or objects are swapped as expected by the baseline method. Our chain of thought approach generated the same outcome. In the fourth, using the baseline method of swap objects, the LLM expert generates a non-sensical outcome. In an example like the last, with the swap attributes directive, the LLM expert chose to abstain. In both these cases, our chain of thought based method generates a good quality hard negative that is meaningful and grammatical while changing few words.

In \cref{tab:diff_gen}, we compare a SCRAMBLe tuned LLaVA-1.5 model (tuned with DPO with synthetic data generated using our chain of thought method) with to a LLaVA-1.5 model and one that we tuned using the baseline data generated with the swap directives. While the baseline data generation approach leads to some improvements on compositionality benchmarks, we find that it degrades performance on long question answering (as measured by MM-Vet performance) immensely. We also found that for long answers it can on occassion degenerate into repeating a single character/phrase in the output. On the other hand, we found that with SCRAMBLe tuning improves performance across the board, including on long-question answering on MM-Vet.

\begin{table}[t]
    \centering
    \scalebox{0.9}{
    \begin{tabular}{@{}lccc@{}}
    \toprule
    & \multicolumn{3}{c}{\textbf{Winoground}} \\
    \textbf{Model Name} & \textbf{Text} & \textbf{Image} & \textbf{Group} \\ \midrule
    Mturk Human$^\S$ & 89.5 & 88.5 & 85.5 \\
    Random Chance$^\S$ & 25.0 & 25.0 & 16.7 \\ \midrule
    \multicolumn{4}{c}{\textbf{Image-Text Encoder Models}} \\
    \midrule
    CLIP (ViT-B/32)$^\S$~\cite{radford2021learning} & 30.8 & 10.5 & 8.0 \\
    SynCLIP~\cite{cascante2023going} & 30.0 & 11.5 & 9.5 \\
    CLIP-SGVL~\cite{herzig2023incorporating} & 32.0 & 14.0 & 9.8 \\
    METER$^\dagger$~\cite{dou2022empirical} & 39.3 & 15.8 & 12.0 \\
    METER+EQSIM~\cite{wang2023equivariant} & 45.0 & 22.8 & 18.8 \\
    FIBER$^\dagger$~\cite{dou2022coarse} & 46.3 & 25.8 & 22.2 \\
    FIBER+EQSIM~\cite{wang2023equivariant} & 51.5 & 31.5 & 27.5 \\
    \midrule
    \multicolumn{4}{c}{\textbf{Proprietary Models + Prompting Methods}} \\
    \midrule
    GPT4V$^\ddagger$~\cite{achiam2023gpt} & 60.3 & 45.3 & 33.5 \\
    GPT4V-CCoT~\cite{mitra2024compositional} & 64.0 & 54.5 & 43.3 \\
    GPT4V+CoCoT~\cite{zhang2024cocot} & 58.5 & 49.5 & 44.5 \\
    \midrule
    \multicolumn{4}{c}{\textbf{Open MLLM Methods + VQAScore Evaluation}} \\
    \midrule
    LLaVA-1.5-13B~\cite{liu2024improved} & 51.5 & 50.5 & 36.5 \\
    CECE (LLaVA-1.6)~\cite{cascante2025natural} & 52.0 & 61.3 & 42.8 \\
    CECE (LLaVA-1.5+LLaVA-1.6)~\cite{cascante2025natural} & 55.0  & 61.3 & 47.5 \\
    Molmo-7B~\cite{deitke2024molmo} & 62.8 & 61.8 & 49.5 \\
    \midrule
    SCRAMBLe-LLaVA-1.5-13B (Ours) & 53.3 & 52.8 & 39.3 \\
    SCRAMBLe-Molmo-7B (Ours) & \textbf{66.8} & \textbf{66.3} & \textbf{54.8} \\
    \bottomrule
    \end{tabular}
    }
    \caption{\textbf{Winoground Performance.} Comparison of SCRAMBLe tuned models with prior work. SCRAMBLe-Molmo improves significantly on previous best. ($^\dagger$reported in \cite{wang2023equivariant}, $^\ddagger$reported in \cite{mitra2024compositional}, $^\S$reported in \cite{thrush2022winoground})}
    \label{tab:winoground_perf}
    \vspace{-2mm}
\end{table}

\noindent\textbf{Winoground Performance.} In \cref{tab:winoground_perf} we compare SCRAMBLe against prior reported compositionality performance on Winoground. In the top section are the performances of a human and a random chance baseline. The next section contains performances of different image and text representation models. CLIP~\cite{radford2021learning} as reported by Thrush~\etal~\cite{thrush2022winoground} performs poorer than random chance, and attempts at improvement by training on compositional data and/or with auxiliary losses~\cite{cascante2023going,herzig2023incorporating} could not improve beyond this. METER~\cite{dou2022empirical} and FIBER~\cite{dou2022coarse} models break past random chance improving with training on additional data with equivariant constraints~\cite{wang2023equivariant}. In the section below, we report performance of GPT-4V~\cite{achiam2023gpt} with different prompting methods~\cite{mitra2024compositional,zhang2024cocot}\footnote{We note that the evaluation methods of these two prompting approaches are different since the former~\cite{mitra2024compositional} restricts itself to models with a single image input while the latter~\cite{zhang2024cocot} uses multiple images in context}. Since GPT-4 is a proprietary model, its output probability distributions have not been made available and thus computing VQAScore~\cite{lin2024evaluating} is not possible. In the last section, we use this evaluation with different methods on open models\footnote{On evaluating LLaVA-1.5-13B with VQAScore, we find that predicting the probability of ``Yes'' as opposed to ``Yes$\langle$\textbackslash s$\rangle$'' leads to better performance (where $\langle$\textbackslash s$\rangle$ is the end of sentence token). We believe this is also more principled since it measures the probability of the model starting its answer with ``Yes'' as opposed to simply giving the one word answer ``Yes.''} and find that SCRAMBLe preference tuning with Molmo-7B-D-0924 leads to much higher performance than previously reported on Winoground.

\begin{figure}[]
    \centering
    \includegraphics[width=\linewidth,trim=0cm 0cm 0cm 0cm,clip]{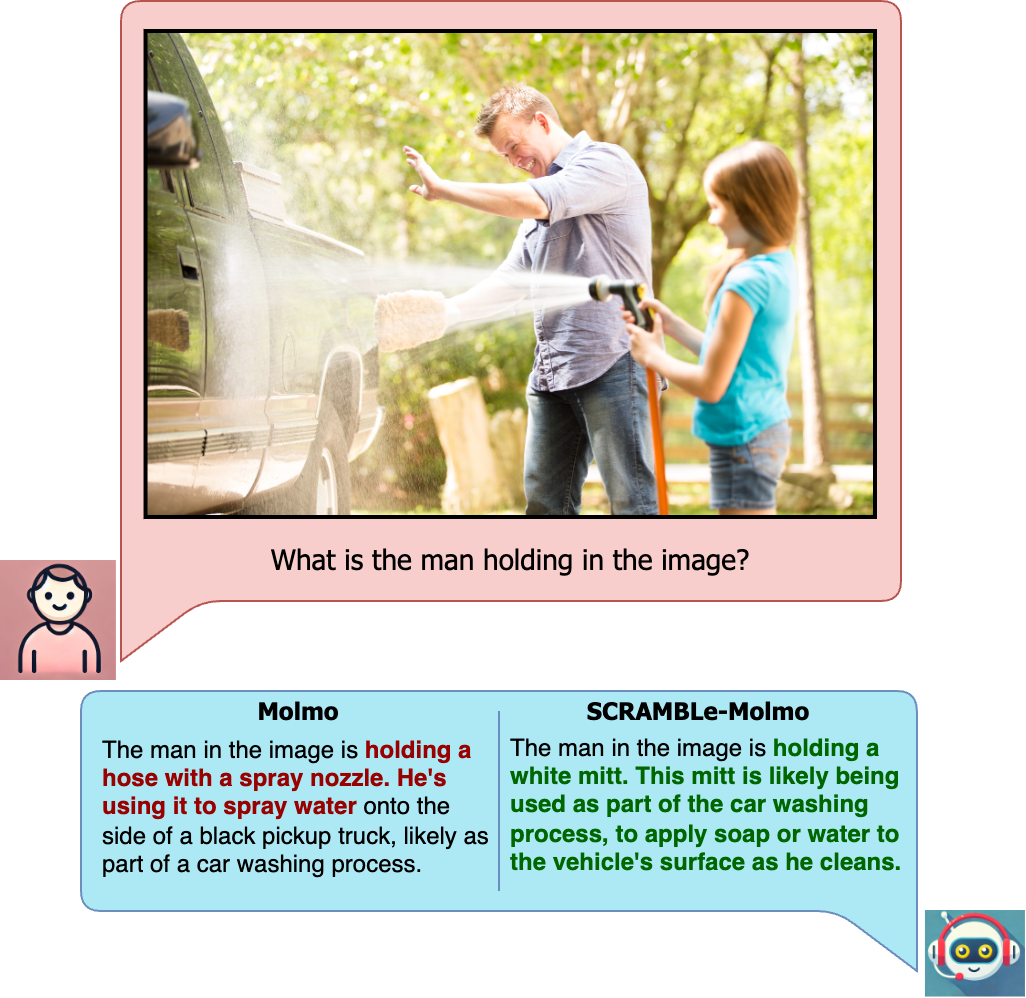}
    \caption{\textbf{Conversing with SCRAMBLe-Molmo (Winoground example).} SCRAMBLe fixes Molmo's mistake identifying what the man holds.}
    \label{fig:chat_eg2}
    \vspace{-2mm}
\end{figure}

\noindent\textbf{Conversing with SCRAMBLe-Molmo.} We show two examples (more in \cref{sec:app-chat_examples}) of chatting with Molmo and SCRAMBLe-Molmo. The first (\cref{fig:chat_eg1}) is an example from EQBen where we find SCRAMBLe-Molmo correctly determines the location of the plate relative to the woman, in contrast to Molmo's mistake. In the second example (\cref{fig:chat_eg2}), we see that SCRAMBLe-Molmo correctly identifies the man is holding a mitt and not a hose, which is what Molmo perceives.

\section{Conclusion}

We introduced SCRAMBLe, a procedure of improving visio-linguistic compositionality of multimodal large language models (MLLMs), by tuning them on synthetic preference data. Our synthetic data uses existing image caption data, along with synthetically generated hard negative captions. The generation procedure is fully automated, using an open-weight LLM expert (Llama-3.1-70B) prompted to transform positive captions into hard negative captions by providing its working in a chain of thought. Tuning state of the art open weight MLLMs with this synthetic preference data significantly improves their compositional reasoning performance while also leading to improvements in general question answering (albeit to a smaller extent).

\section*{Acknowledgements} 
Venkatesh Saligrama and Samarth Mishra were supported by the Army Research Office Grant W911NF2110246, AFRLGrant FA8650-22-C1039, the National Science Foundation grants CPS-2317079, CCF2007350 and CCF-1955981. Kate Saenko was supported by the National Science Foundation.

{
    \small
    \bibliographystyle{ieeenat_fullname}
    \bibliography{main}

\begin{thebibliography}{60}
\providecommand{\natexlab}[1]{#1}
\providecommand{\url}[1]{\texttt{#1}}
\expandafter\ifx\csname urlstyle\endcsname\relax
  \providecommand{\doi}[1]{doi: #1}\else
  \providecommand{\doi}{doi: \begingroup \urlstyle{rm}\Url}\fi

\bibitem[Achiam et~al.(2023)Achiam, Adler, Agarwal, Ahmad, Akkaya, Aleman, Almeida, Altenschmidt, Altman, Anadkat, et~al.]{achiam2023gpt}
Josh Achiam, Steven Adler, Sandhini Agarwal, Lama Ahmad, Ilge Akkaya, Florencia~Leoni Aleman, Diogo Almeida, Janko Altenschmidt, Sam Altman, Shyamal Anadkat, et~al.
\newblock Gpt-4 technical report.
\newblock \emph{arXiv preprint arXiv:2303.08774}, 2023.

\bibitem[Agrawal et~al.(2024)Agrawal, Antoniak, Hanna, Chaplot, Chudnovsky, Garg, Gervet, Ghosh, H{\'e}liou, Jacob, et~al.]{agrawal2024pixtral}
Pravesh Agrawal, Szymon Antoniak, Emma~Bou Hanna, Devendra Chaplot, Jessica Chudnovsky, Saurabh Garg, Theophile Gervet, Soham Ghosh, Am{\'e}lie H{\'e}liou, Paul Jacob, et~al.
\newblock Pixtral 12b.
\newblock \emph{arXiv preprint arXiv:2410.07073}, 2024.

\bibitem[Anthropic(2024)]{anthropic2024introducing}
AI Anthropic.
\newblock Introducing the next generation of claude, 2024.

\bibitem[Bommasani et~al.(2021)Bommasani, Hudson, Adeli, Altman, Arora, von Arx, Bernstein, Bohg, Bosselut, Brunskill, et~al.]{bommasani2021opportunities}
Rishi Bommasani, Drew~A Hudson, Ehsan Adeli, Russ Altman, Simran Arora, Sydney von Arx, Michael~S Bernstein, Jeannette Bohg, Antoine Bosselut, Emma Brunskill, et~al.
\newblock On the opportunities and risks of foundation models.
\newblock \emph{arXiv preprint arXiv:2108.07258}, 2021.

\bibitem[Cascante-Bonilla et~al.(2023)Cascante-Bonilla, Shehada, Smith, Doveh, Kim, Panda, Varol, Oliva, Ordonez, Feris, et~al.]{cascante2023going}
Paola Cascante-Bonilla, Khaled Shehada, James~Seale Smith, Sivan Doveh, Donghyun Kim, Rameswar Panda, Gul Varol, Aude Oliva, Vicente Ordonez, Rogerio Feris, et~al.
\newblock Going beyond nouns with vision \& language models using synthetic data.
\newblock In \emph{Proceedings of the IEEE/CVF International Conference on Computer Vision}, pages 20155--20165, 2023.

\bibitem[Cascante-Bonilla et~al.(2025)Cascante-Bonilla, Hou, Cao, Daum{\'e}~III, and Rudinger]{cascante2025natural}
Paola Cascante-Bonilla, Yu Hou, Yang~Trista Cao, Hal Daum{\'e}~III, and Rachel Rudinger.
\newblock Natural language inference improves compositionality in vision-language models.
\newblock In \emph{International Conference on Learning Representations (ICLR)}, 2025.

\bibitem[Chen et~al.(2020)Chen, Li, Yu, El~Kholy, Ahmed, Gan, Cheng, and Liu]{chen2020uniter}
Yen-Chun Chen, Linjie Li, Licheng Yu, Ahmed El~Kholy, Faisal Ahmed, Zhe Gan, Yu Cheng, and Jingjing Liu.
\newblock Uniter: Universal image-text representation learning.
\newblock In \emph{European conference on computer vision}, pages 104--120. Springer, 2020.

\bibitem[Chiang et~al.(2023)Chiang, Li, Lin, Sheng, Wu, Zhang, Zheng, Zhuang, Zhuang, Gonzalez, et~al.]{chiang2023vicuna}
Wei-Lin Chiang, Zhuohan Li, Zi Lin, Ying Sheng, Zhanghao Wu, Hao Zhang, Lianmin Zheng, Siyuan Zhuang, Yonghao Zhuang, Joseph~E Gonzalez, et~al.
\newblock Vicuna: An open-source chatbot impressing gpt-4 with 90\%* chatgpt quality.
\newblock \emph{See https://vicuna. lmsys. org (accessed 14 April 2023)}, 2\penalty0 (3):\penalty0 6, 2023.

\bibitem[Deitke et~al.(2024)Deitke, Clark, Lee, Tripathi, Yang, Park, Salehi, Muennighoff, Lo, Soldaini, et~al.]{deitke2024molmo}
Matt Deitke, Christopher Clark, Sangho Lee, Rohun Tripathi, Yue Yang, Jae~Sung Park, Mohammadreza Salehi, Niklas Muennighoff, Kyle Lo, Luca Soldaini, et~al.
\newblock Molmo and pixmo: Open weights and open data for state-of-the-art multimodal models.
\newblock \emph{arXiv preprint arXiv:2409.17146}, 2024.

\bibitem[Diwan et~al.(2022)Diwan, Berry, Choi, Harwath, and Mahowald]{diwan2022winoground}
Anuj Diwan, Layne Berry, Eunsol Choi, David Harwath, and Kyle Mahowald.
\newblock Why is winoground hard? investigating failures in visuolinguistic compositionality.
\newblock In \emph{Proceedings of the 60th Annual Meeting of the Association for Computational Linguistics (ACL)}, Dublin, Ireland, 2022. Association for Computational Linguistics.

\bibitem[Dou et~al.(2022{\natexlab{a}})Dou, Kamath, Gan, Zhang, Wang, Li, Liu, Liu, LeCun, Peng, et~al.]{dou2022coarse}
Zi-Yi Dou, Aishwarya Kamath, Zhe Gan, Pengchuan Zhang, Jianfeng Wang, Linjie Li, Zicheng Liu, Ce Liu, Yann LeCun, Nanyun Peng, et~al.
\newblock Coarse-to-fine vision-language pre-training with fusion in the backbone.
\newblock \emph{Advances in neural information processing systems}, 35:\penalty0 32942--32956, 2022{\natexlab{a}}.

\bibitem[Dou et~al.(2022{\natexlab{b}})Dou, Xu, Gan, Wang, Wang, Wang, Zhu, Zhang, Yuan, Peng, et~al.]{dou2022empirical}
Zi-Yi Dou, Yichong Xu, Zhe Gan, Jianfeng Wang, Shuohang Wang, Lijuan Wang, Chenguang Zhu, Pengchuan Zhang, Lu Yuan, Nanyun Peng, et~al.
\newblock An empirical study of training end-to-end vision-and-language transformers.
\newblock In \emph{Proceedings of the IEEE/CVF Conference on Computer Vision and Pattern Recognition}, pages 18166--18176, 2022{\natexlab{b}}.

\bibitem[Doveh et~al.(2023{\natexlab{a}})Doveh, Arbelle, Harary, Herzig, Kim, Cascante-Bonilla, Alfassy, Panda, Giryes, Feris, et~al.]{doveh2023dense}
Sivan Doveh, Assaf Arbelle, Sivan Harary, Roei Herzig, Donghyun Kim, Paola Cascante-Bonilla, Amit Alfassy, Rameswar Panda, Raja Giryes, Rogerio Feris, et~al.
\newblock Dense and aligned captions (dac) promote compositional reasoning in vl models.
\newblock \emph{Advances in Neural Information Processing Systems}, 36:\penalty0 76137--76150, 2023{\natexlab{a}}.

\bibitem[Doveh et~al.(2023{\natexlab{b}})Doveh, Arbelle, Harary, Schwartz, Herzig, Giryes, Feris, Panda, Ullman, and Karlinsky]{doveh2023teaching}
Sivan Doveh, Assaf Arbelle, Sivan Harary, Eli Schwartz, Roei Herzig, Raja Giryes, Rogerio Feris, Rameswar Panda, Shimon Ullman, and Leonid Karlinsky.
\newblock Teaching structured vision \& language concepts to vision \& language models.
\newblock In \emph{Proceedings of the IEEE/CVF Conference on Computer Vision and Pattern Recognition}, pages 2657--2668, 2023{\natexlab{b}}.

\bibitem[Dubey et~al.(2024)Dubey, Jauhri, Pandey, Kadian, Al-Dahle, Letman, Mathur, Schelten, Yang, Fan, et~al.]{dubey2024llama}
Abhimanyu Dubey, Abhinav Jauhri, Abhinav Pandey, Abhishek Kadian, Ahmad Al-Dahle, Aiesha Letman, Akhil Mathur, Alan Schelten, Amy Yang, Angela Fan, et~al.
\newblock The llama 3 herd of models.
\newblock \emph{arXiv preprint arXiv:2407.21783}, 2024.

\bibitem[Face(2024)]{huggingfacehub}
Hugging Face.
\newblock Hugging face hub: A platform for sharing machine learning models, datasets, and spaces.
\newblock \url{https://huggingface.co/docs/hub/en/index}, 2024.
\newblock Accessed: YYYY-MM-DD.

\bibitem[Herzig et~al.(2023)Herzig, Mendelson, Karlinsky, Arbelle, Feris, Darrell, and Globerson]{herzig2023incorporating}
Roei Herzig, Alon Mendelson, Leonid Karlinsky, Assaf Arbelle, Rogerio Feris, Trevor Darrell, and Amir Globerson.
\newblock Incorporating structured representations into pretrained vision \& language models using scene graphs.
\newblock In \emph{Proceedings of the 2023 Conference on Empirical Methods in Natural Language Processing (EMNLP)}, Singapore, 2023. Association for Computational Linguistics.

\bibitem[Hessel et~al.(2021)Hessel, Holtzman, Forbes, Bras, and Choi]{hessel2021clipscore}
Jack Hessel, Ari Holtzman, Maxwell Forbes, Ronan~Le Bras, and Yejin Choi.
\newblock Clipscore: A reference-free evaluation metric for image captioning.
\newblock In \emph{Proceedings of the 2021 Conference on Empirical Methods in Natural Language Processing (EMNLP)}, 2021.

\bibitem[Honnibal and Montani(2015)]{spacy}
Matthew Honnibal and Ines Montani.
\newblock spacy: Industrial-strength natural language processing in python, 2015.

\bibitem[Hsieh et~al.(2024)Hsieh, Zhang, Ma, Kembhavi, and Krishna]{hsieh2024sugarcrepe}
Cheng-Yu Hsieh, Jieyu Zhang, Zixian Ma, Aniruddha Kembhavi, and Ranjay Krishna.
\newblock Sugarcrepe: Fixing hackable benchmarks for vision-language compositionality.
\newblock \emph{Advances in neural information processing systems}, 36, 2024.

\bibitem[Hu et~al.(2022)Hu, Shen, Wallis, Allen-Zhu, Li, Wang, Wang, and Chen]{hu2021lora}
Edward~J. Hu, Yelong Shen, Phillip Wallis, Zeyuan Allen-Zhu, Yuanzhi Li, Shean Wang, Lu Wang, and Weizhu Chen.
\newblock Lora: Low-rank adaptation of large language models.
\newblock In \emph{Proceedings of the International Conference on Learning Representations (ICLR)}, 2022.

\bibitem[Huang et~al.(2024)Huang, Lin, Mirza, Hansen, Doveh, Butoi, Herzig, Arbelle, Kuehne, Darrell, Gan, Oliva, Feris, and Karlinsky]{huang2024conme}
Irene Huang, Wei Lin, Muhammad~Jehanzeb Mirza, Jacob~A. Hansen, Sivan Doveh, Victor~Ion Butoi, Roei Herzig, Assaf Arbelle, Hilde Kuehne, Trevor Darrell, Chuang Gan, Aude Oliva, Rogerio Feris, and Leonid Karlinsky.
\newblock Conme: Rethinking evaluation of compositional reasoning for modern vlms.
\newblock In \emph{Advances in Neural Information Processing Systems 37 (NeurIPS 2024), Datasets and Benchmarks Track}, 2024.

\bibitem[Kreutzer et~al.(2018)Kreutzer, Uyheng, and Riezler]{kreutzer2018reliability}
Julia Kreutzer, Joshua Uyheng, and Stefan Riezler.
\newblock Reliability and learnability of human bandit feedback for sequence-to-sequence reinforcement learning.
\newblock In \emph{Proceedings of the 56th Annual Meeting of the Association for Computational Linguistics (ACL)}, Melbourne, Australia, 2018. Association for Computational Linguistics.

\bibitem[Lake et~al.(2017)Lake, Ullman, Tenenbaum, and Gershman]{lake2017building}
Brenden~M Lake, Tomer~D Ullman, Joshua~B Tenenbaum, and Samuel~J Gershman.
\newblock Building machines that learn and think like people.
\newblock \emph{Behavioral and brain sciences}, 40:\penalty0 e253, 2017.

\bibitem[Li et~al.(2023)Li, Wang, Wang, Ge, Ge, and Shan]{li2023seed}
Bohao Li, Rui Wang, Guangzhi Wang, Yuying Ge, Yixiao Ge, and Ying Shan.
\newblock Seed-bench: Benchmarking multimodal llms with generative comprehension.
\newblock \emph{arXiv preprint arXiv:2307.16125}, 2023.

\bibitem[Lin et~al.(2014)Lin, Maire, Belongie, Hays, Perona, Ramanan, Doll{\'a}r, and Zitnick]{lin2014microsoft}
Tsung-Yi Lin, Michael Maire, Serge Belongie, James Hays, Pietro Perona, Deva Ramanan, Piotr Doll{\'a}r, and C~Lawrence Zitnick.
\newblock Microsoft coco: Common objects in context.
\newblock In \emph{Computer Vision--ECCV 2014: 13th European Conference, Zurich, Switzerland, September 6-12, 2014, Proceedings, Part V 13}, pages 740--755. Springer, 2014.

\bibitem[Lin et~al.(2024)Lin, Pathak, Li, Li, Xia, Neubig, Zhang, and Ramanan]{lin2024evaluating}
Zhiqiu Lin, Deepak Pathak, Baiqi Li, Jiayao Li, Xide Xia, Graham Neubig, Pengchuan Zhang, and Deva Ramanan.
\newblock Evaluating text-to-visual generation with image-to-text generation.
\newblock \emph{arXiv preprint arXiv:2404.01291}, 2024.

\bibitem[Liu et~al.(2024{\natexlab{a}})Liu, Li, Li, and Lee]{liu2024improved}
Haotian Liu, Chunyuan Li, Yuheng Li, and Yong~Jae Lee.
\newblock Improved baselines with visual instruction tuning.
\newblock In \emph{Proceedings of the IEEE/CVF Conference on Computer Vision and Pattern Recognition}, pages 26296--26306, 2024{\natexlab{a}}.

\bibitem[Liu et~al.(2024{\natexlab{b}})Liu, Li, Wu, and Lee]{liu2024visual}
Haotian Liu, Chunyuan Li, Qingyang Wu, and Yong~Jae Lee.
\newblock Visual instruction tuning.
\newblock \emph{Advances in neural information processing systems}, 36, 2024{\natexlab{b}}.

\bibitem[Liu et~al.(2023)Liu, Wang, Wang, Smith, Choi, and Hajishirzi]{liu2023vera}
Jiacheng Liu, Wenya Wang, Dianzhuo Wang, Noah~A. Smith, Yejin Choi, and Hannaneh Hajishirzi.
\newblock {Vera: A General-Purpose Plausibility Estimation Model for Commonsense Statements}.
\newblock In \emph{Proceedings of the 2023 Conference on Empirical Methods in Natural Language Processing (EMNLP)}, Singapore, 2023. Association for Computational Linguistics.

\bibitem[Liu et~al.(2024{\natexlab{c}})Liu, Yu, Lin, Pathak, and Ramanan]{liu2024language}
Shihong Liu, Samuel Yu, Zhiqiu Lin, Deepak Pathak, and Deva Ramanan.
\newblock Language models as black-box optimizers for vision-language models.
\newblock In \emph{Proceedings of the IEEE/CVF Conference on Computer Vision and Pattern Recognition}, pages 12687--12697, 2024{\natexlab{c}}.

\bibitem[Ma et~al.(2023)Ma, Hong, Gul, Gandhi, Gao, and Krishna]{ma2023crepe}
Zixian Ma, Jerry Hong, Mustafa~Omer Gul, Mona Gandhi, Irena Gao, and Ranjay Krishna.
\newblock Crepe: Can vision-language foundation models reason compositionally?
\newblock In \emph{Proceedings of the IEEE/CVF Conference on Computer Vision and Pattern Recognition}, pages 10910--10921, 2023.

\bibitem[Mitra et~al.(2024)Mitra, Huang, Darrell, and Herzig]{mitra2024compositional}
Chancharik Mitra, Brandon Huang, Trevor Darrell, and Roei Herzig.
\newblock Compositional chain-of-thought prompting for large multimodal models.
\newblock In \emph{Proceedings of the IEEE/CVF Conference on Computer Vision and Pattern Recognition}, pages 14420--14431, 2024.

\bibitem[Morris et~al.(2020)Morris, Lifland, Yoo, Grigsby, Jin, and Qi]{morris2020textattack}
John~X. Morris, Eli Lifland, Jin~Yong Yoo, Jake Grigsby, Di Jin, and Yanjun Qi.
\newblock Textattack: A framework for adversarial attacks, data augmentation, and adversarial training in nlp.
\newblock In \emph{Proceedings of the 2020 Conference on Empirical Methods in Natural Language Processing (EMNLP)}, pages 119--139, Online, 2020. Association for Computational Linguistics.

\bibitem[{OpenAI}(2024)]{openai2024gpt4o}
{OpenAI}.
\newblock Hello gpt-4o, 2024.
\newblock Accessed: 2024-11-06.

\bibitem[Ouyang et~al.(2022)Ouyang, Wu, Jiang, Almeida, Wainwright, Mishkin, Zhang, Agarwal, Slama, Ray, et~al.]{ouyang2022training}
Long Ouyang, Jeffrey Wu, Xu Jiang, Diogo Almeida, Carroll Wainwright, Pamela Mishkin, Chong Zhang, Sandhini Agarwal, Katarina Slama, Alex Ray, et~al.
\newblock Training language models to follow instructions with human feedback.
\newblock \emph{Advances in neural information processing systems}, 35:\penalty0 27730--27744, 2022.

\bibitem[Paszke et~al.(2019)Paszke, Gross, Massa, Lerer, Bradbury, Chanan, Killeen, Lin, Gimelshein, Antiga, et~al.]{paszke2019pytorch}
Adam Paszke, Sam Gross, Francisco Massa, Adam Lerer, James Bradbury, Gregory Chanan, Trevor Killeen, Zeming Lin, Natalia Gimelshein, Luca Antiga, et~al.
\newblock Pytorch: An imperative style, high-performance deep learning library.
\newblock \emph{Advances in neural information processing systems}, 32, 2019.

\bibitem[Pratt et~al.(2023)Pratt, Covert, Liu, and Farhadi]{pratt2023does}
Sarah Pratt, Ian Covert, Rosanne Liu, and Ali Farhadi.
\newblock What does a platypus look like? generating customized prompts for zero-shot image classification.
\newblock In \emph{Proceedings of the IEEE/CVF International Conference on Computer Vision}, pages 15691--15701, 2023.

\bibitem[Radford et~al.(2021)Radford, Kim, Hallacy, Ramesh, Goh, Agarwal, Sastry, Askell, Mishkin, Clark, et~al.]{radford2021learning}
Alec Radford, Jong~Wook Kim, Chris Hallacy, Aditya Ramesh, Gabriel Goh, Sandhini Agarwal, Girish Sastry, Amanda Askell, Pamela Mishkin, Jack Clark, et~al.
\newblock Learning transferable visual models from natural language supervision.
\newblock In \emph{International conference on machine learning}, pages 8748--8763. PMLR, 2021.

\bibitem[Rafailov et~al.(2024)Rafailov, Sharma, Mitchell, Manning, Ermon, and Finn]{rafailov2024direct}
Rafael Rafailov, Archit Sharma, Eric Mitchell, Christopher~D Manning, Stefano Ermon, and Chelsea Finn.
\newblock Direct preference optimization: Your language model is secretly a reward model.
\newblock \emph{Advances in Neural Information Processing Systems}, 36, 2024.

\bibitem[Ramesh et~al.(2022)Ramesh, Dhariwal, Nichol, Chu, and Chen]{ramesh2022hierarchical}
Aditya Ramesh, Prafulla Dhariwal, Alex Nichol, Casey Chu, and Mark Chen.
\newblock Hierarchical text-conditional image generation with clip latents.
\newblock \emph{arXiv preprint arXiv:2204.06125}, 1\penalty0 (2):\penalty0 3, 2022.

\bibitem[Ray et~al.(2024)Ray, Radenovic, Dubey, Plummer, Krishna, and Saenko]{ray2024cola}
Arijit Ray, Filip Radenovic, Abhimanyu Dubey, Bryan Plummer, Ranjay Krishna, and Kate Saenko.
\newblock Cola: A benchmark for compositional text-to-image retrieval.
\newblock \emph{Advances in Neural Information Processing Systems}, 36, 2024.

\bibitem[Rombach et~al.(2022)Rombach, Blattmann, Lorenz, Esser, and Ommer]{rombach2022high}
Robin Rombach, Andreas Blattmann, Dominik Lorenz, Patrick Esser, and Bj{\"o}rn Ommer.
\newblock High-resolution image synthesis with latent diffusion models.
\newblock In \emph{Proceedings of the IEEE/CVF conference on computer vision and pattern recognition}, pages 10684--10695, 2022.

\bibitem[Saito et~al.(2023)Saito, Sohn, Zhang, Li, Lee, Saenko, and Pfister]{saito2023pic2word}
Kuniaki Saito, Kihyuk Sohn, Xiang Zhang, Chun-Liang Li, Chen-Yu Lee, Kate Saenko, and Tomas Pfister.
\newblock Pic2word: Mapping pictures to words for zero-shot composed image retrieval.
\newblock In \emph{Proceedings of the IEEE/CVF Conference on Computer Vision and Pattern Recognition}, pages 19305--19314, 2023.

\bibitem[Stiennon et~al.(2020)Stiennon, Ouyang, Wu, Ziegler, Lowe, Voss, Radford, Amodei, and Christiano]{stiennon2020learning}
Nisan Stiennon, Long Ouyang, Jeffrey Wu, Daniel Ziegler, Ryan Lowe, Chelsea Voss, Alec Radford, Dario Amodei, and Paul~F Christiano.
\newblock Learning to summarize with human feedback.
\newblock \emph{Advances in Neural Information Processing Systems}, 33:\penalty0 3008--3021, 2020.

\bibitem[Thrush et~al.(2022)Thrush, Jiang, Bartolo, Singh, Williams, Kiela, and Ross]{thrush2022winoground}
Tristan Thrush, Ryan Jiang, Max Bartolo, Amanpreet Singh, Adina Williams, Douwe Kiela, and Candace Ross.
\newblock Winoground: Probing vision and language models for visio-linguistic compositionality.
\newblock In \emph{Proceedings of the IEEE/CVF Conference on Computer Vision and Pattern Recognition}, pages 5238--5248, 2022.

\bibitem[Touvron et~al.(2023)Touvron, Martin, Stone, Albert, Almahairi, Babaei, Bashlykov, Batra, Bhargava, Bhosale, et~al.]{touvron2023llama}
Hugo Touvron, Louis Martin, Kevin Stone, Peter Albert, Amjad Almahairi, Yasmine Babaei, Nikolay Bashlykov, Soumya Batra, Prajjwal Bhargava, Shruti Bhosale, et~al.
\newblock Llama 2: Open foundation and fine-tuned chat models.
\newblock \emph{arXiv preprint arXiv:2307.09288}, 2023.

\bibitem[von Werra et~al.(2020)von Werra, Belkada, Tunstall, Beeching, Thrush, Lambert, Huang, Rasul, and Gallouédec]{vonwerra2022trl}
Leandro von Werra, Younes Belkada, Lewis Tunstall, Edward Beeching, Tristan Thrush, Nathan Lambert, Shengyi Huang, Kashif Rasul, and Quentin Gallouédec.
\newblock Trl: Transformer reinforcement learning.
\newblock \url{https://github.com/huggingface/trl}, 2020.

\bibitem[Wang et~al.(2023)Wang, Lin, Li, Lin, Yang, Zhang, Liu, and Wang]{wang2023equivariant}
Tan Wang, Kevin Lin, Linjie Li, Chung-Ching Lin, Zhengyuan Yang, Hanwang Zhang, Zicheng Liu, and Lijuan Wang.
\newblock Equivariant similarity for vision-language foundation models.
\newblock In \emph{Proceedings of the IEEE/CVF International Conference on Computer Vision}, pages 11998--12008, 2023.

\bibitem[Wei et~al.(2022)Wei, Wang, Schuurmans, Bosma, Xia, Chi, Le, Zhou, et~al.]{wei2022chain}
Jason Wei, Xuezhi Wang, Dale Schuurmans, Maarten Bosma, Fei Xia, Ed Chi, Quoc~V Le, Denny Zhou, et~al.
\newblock Chain-of-thought prompting elicits reasoning in large language models.
\newblock \emph{Advances in neural information processing systems}, 35:\penalty0 24824--24837, 2022.

\bibitem[Wolf et~al.(2020)Wolf, Debut, Sanh, Chaumond, Delangue, Moi, Cistac, Rault, Louf, Funtowicz, Davison, Shleifer, von Platen, Ma, Jernite, Plu, Xu, Scao, Gugger, Drame, Lhoest, and Rush]{wolf-etal-2020-transformers}
Thomas Wolf, Lysandre Debut, Victor Sanh, Julien Chaumond, Clement Delangue, Anthony Moi, Pierric Cistac, Tim Rault, Rémi Louf, Morgan Funtowicz, Joe Davison, Sam Shleifer, Patrick von Platen, Clara Ma, Yacine Jernite, Julien Plu, Canwen Xu, Teven~Le Scao, Sylvain Gugger, Mariama Drame, Quentin Lhoest, and Alexander~M. Rush.
\newblock Transformers: State-of-the-art natural language processing.
\newblock In \emph{Proceedings of the 2020 Conference on Empirical Methods in Natural Language Processing: System Demonstrations}, pages 38--45, Online, 2020. Association for Computational Linguistics.

\bibitem[Yu et~al.(2023)Yu, Yang, Li, Wang, Lin, Liu, Wang, and Wang]{yu2023mm}
Weihao Yu, Zhengyuan Yang, Linjie Li, Jianfeng Wang, Kevin Lin, Zicheng Liu, Xinchao Wang, and Lijuan Wang.
\newblock Mm-vet: Evaluating large multimodal models for integrated capabilities.
\newblock \emph{arXiv preprint arXiv:2308.02490}, 2023.

\bibitem[Yuksekgonul et~al.(2022)Yuksekgonul, Bianchi, Kalluri, Jurafsky, and Zou]{yuksekgonul2022and}
Mert Yuksekgonul, Federico Bianchi, Pratyusha Kalluri, Dan Jurafsky, and James Zou.
\newblock When and why vision-language models behave like bags-of-words, and what to do about it?
\newblock \emph{arXiv preprint arXiv:2210.01936}, 2022.

\bibitem[Zhang et~al.(2024)Zhang, Yang, Lyu, Jin, Yao, Chen, and Luo]{zhang2024cocot}
Daoan Zhang, Junming Yang, Hanjia Lyu, Zijian Jin, Yuan Yao, Mingkai Chen, and Jiebo Luo.
\newblock Cocot: Contrastive chain-of-thought prompting for large multimodal models with multiple image inputs.
\newblock \emph{arXiv preprint arXiv:2401.02582}, 2024.

\bibitem[Zhang et~al.(2021)Zhang, Li, Hu, Yang, Zhang, Wang, Choi, and Gao]{zhang2021vinvl}
Pengchuan Zhang, Xiujun Li, Xiaowei Hu, Jianwei Yang, Lei Zhang, Lijuan Wang, Yejin Choi, and Jianfeng Gao.
\newblock Vinvl: Revisiting visual representations in vision-language models.
\newblock In \emph{Proceedings of the IEEE/CVF conference on computer vision and pattern recognition}, pages 5579--5588, 2021.

\bibitem[Zhao et~al.(2022)Zhao, Zhang, Zhu, Shen, Lee, Lu, and Yin]{zhao2022explainable}
Tiancheng Zhao, Tianqi Zhang, Mingwei Zhu, Haozhan Shen, Kyusong Lee, Xiaopeng Lu, and Jianwei Yin.
\newblock An explainable toolbox for evaluating pre-trained vision-language models.
\newblock In \emph{Proceedings of the 2022 Conference on Empirical Methods in Natural Language Processing: System Demonstrations}, pages 30--37, Abu Dhabi, UAE, 2022. Association for Computational Linguistics.

\bibitem[Zhou et~al.(2022)Zhou, Yang, Loy, and Liu]{zhou2022learning}
Kaiyang Zhou, Jingkang Yang, Chen~Change Loy, and Ziwei Liu.
\newblock Learning to prompt for vision-language models.
\newblock \emph{International Journal of Computer Vision}, 130\penalty0 (9):\penalty0 2337--2348, 2022.

\bibitem[Zhou et~al.(2024)Zhou, Cui, Rafailov, Finn, and Yao]{zhou2024aligning}
Yiyang Zhou, Chenhang Cui, Rafael Rafailov, Chelsea Finn, and Huaxiu Yao.
\newblock Aligning modalities in vision large language models via preference fine-tuning, 2024.

\bibitem[Zhu et~al.(2023)Zhu, Chen, Shen, Li, and Elhoseiny]{zhu2023minigpt}
Deyao Zhu, Jun Chen, Xiaoqian Shen, Xiang Li, and Mohamed Elhoseiny.
\newblock Minigpt-4: Enhancing vision-language understanding with advanced large language models.
\newblock \emph{arXiv preprint arXiv:2304.10592}, 2023.

\bibitem[Ziegler et~al.(2019)Ziegler, Stiennon, Wu, Brown, Radford, Amodei, Christiano, and Irving]{ziegler2019fine}
Daniel~M Ziegler, Nisan Stiennon, Jeffrey Wu, Tom~B Brown, Alec Radford, Dario Amodei, Paul Christiano, and Geoffrey Irving.
\newblock Fine-tuning language models from human preferences.
\newblock \emph{arXiv preprint arXiv:1909.08593}, 2019.

\end{thebibliography}
}

\break
\section*{Appendices}
\appendix

\section{Alternative Data Generation Approach : Feedback Loop}

We experimented with a third method of synthetic data generation but found the data quality to be poorer than that generated using chain of thought. For completeness, we report the method and experiments here. In this approach, we provide feedback to the LLM expert in context in an attempt to get it to refine the negative caption that it generated. We provide feedback along 4 different dimensions : 

\begin{itemize}
    \item \textbf{Plausibility} : We use the Vera model~\cite{liu2023vera} to score how plausible a generated caption is, from 0 to 1. An illogical/nonsensical caption would have a low score, while a caption that is plausible would score higher.
    \item \textbf{Grammar} : We use the grammar model from TextAttack~\cite{morris2020textattack} to score how grammatical a generated caption is, from 0 to 1. Lower scores indicate poorer grammar.
    \item \textbf{Distinction} : This is a binary response which is 1/Yes if the generated caption is visually distinct from the original caption, and 0/No otherwise. We use a different Llama-3.1 expert that determines this given the original caption and the generated caption.
    \item \textbf{New and Missing Words} : We lemmatize all the words in the original and the new caption using spaCy~\cite{spacy} and in the feedback mention which extra words the LLM has used and which words from the original caption are missing in the generated caption.
\end{itemize}

The start of an example conversation is in \cref{fig:feedback-prompt}. Since there is no one measure of the quality of a hard negative caption, we attempt to get the LLM to optimize each of the scores the best it can. In this setup, the LLM thus acts as a black-box optimizer~\cite{liu2024language}.

The initial prompt to the LLM expert mentions the same requirements as in the Swap-Objects prompt (\cref{fig:swap-obj-prompt-supp}), with the distinction of asking the expert to rearrange words in the input caption as it sees fit. After the LLM generates a caption, we run the feedback models and provide the scores to the expert in the same context. We repeat this process for 5 iterations and pick the caption which is judged distinct from the input and has the highest average score over plausibility, grammar and word similarity (proportional to the Jaccard similarity between the sets of words in the two captions : a score of 1 indicating the same set of words were used in the two captions). A full example conversation is in \cref{sec:feedback_convo}.

\begin{figure}
    \centering
    \fcolorbox{black}{gray!10}{%
    \parbox{0.95\linewidth}{%
        \ttfamily
        \fontsize{8}{11}\selectfont
        Given an input caption describing a scene, your task is to rearrange words in it to make a new caption. The new caption must meet the following three requirements:
        \begin{enumerate}
            \item It must describe a scene with visual differences to the input caption.
            \item It must be fluent and grammatically correct.
            \item It must make logical sense.
        \end{enumerate}

        You can choose to abstain and output 'NA' if it is not possible to generate a negative caption for the given input.

        To help with your task, I will rate your output based on grammar (0-1), plausibility (0-1), and whether there are visual differences between the original caption and your output (Yes/No).

        Here is the input caption: A white horse pulling a cart down a street.

        \textcolor[rgb]{0, 0.5, 0}{New Caption: A cart is being pulled down the street by a white horse.}

        FEEDBACK: \\
        Your grammar score is \textcolor[rgb]{0, 0, 0.5}{0.99}. \\
        Your plausibility score is \textcolor[rgb]{0, 0, 0.5}{0.56}. \\
        Is the output caption visually different from the original caption? : \textcolor[rgb]{0, 0, 0.5}{No} \\
        Your output caption has extra words (lemmatized): \textcolor[rgb]{0, 0, 0.5}{'the', 'by', 'be'}. \\
        Can you please try again?\\
        \vdots
        }%
    }
    \caption{\textbf{Feedback Loop.} Example prompt to generate a negative caption (generated caption in green) and feedback generated using auxiliary models (in blue). The Llama-3.1 expert is provided the feedback and prompted to try again for 5 iterations.}
    \label{fig:feedback-prompt}
    \vspace{-3mm}
\end{figure}

\begin{table*}[t]
    \centering
    \scalebox{0.9}{
    \begin{tabular}{@{}ll|cccc|cc@{}}
    \toprule
    & & \multicolumn{4}{c|}{\textbf{Compositionality Benchmarks}} & \multicolumn{2}{c}{\textbf{Control Benchmarks}} \\ \midrule
    \multicolumn{1}{c}{\textbf{Model   Name}} & \multicolumn{1}{c|}{\textbf{Tuning Data}} & \multicolumn{1}{c}{\textbf{Winoground}} & \multicolumn{1}{c}{\textbf{EqBen}} & \multicolumn{1}{c}{\textbf{COLA}} & \multicolumn{1}{c|}{\textbf{ConMe}} & \textbf{SEED-Bench} & \multicolumn{1}{c}{\textbf{MM-Vet}} \\ \midrule
    LLaVA-1.5-13B & \multicolumn{1}{c|}{-} & 36.5 & 36.4 & 49.5 & 62.3 & 68.23 & 36.2 $\pm$ 0.3 \\
    Baseline (w tuning) & Swap Obj/Att & 38.8 & 36.4 & 52.9 & 64.4 & \textbf{68.49} & 30.7 $\pm$ 0.4 \\
    Baseline-II (w tuning) & Feedback Loop & 37.5 & 33.6 & \textbf{57.1} & \textbf{65.4} & 67.78 & 36.3 $\pm$ 0.2 \\
    SCRAMBLe (Ours) & Chain of Thought & \textbf{39.3} & \textbf{39.3} & 55.7 & 64.5 & 68.19 & \textbf{38.6 $\pm$ 0.1} \\
    \bottomrule
    \end{tabular}
    }
    \caption{\textbf{SCRAMBLe vs other caption generation methods.} Adding to Tab 4 from the main paper, we report the results of tuning the base LLaVA-1.5-13B with synthetic data from the synthetic data generated using a feedback loop. We find that this method does well on some compositionality benchmarks(COLA and ConMe) but is not consistently better than the base LLaVA model especially on the control benchmarks.}
    \label{tab:diff_gen_supp}
    \vspace{-2mm}
\end{table*}

\noindent\textbf{Results.} The results of tuning a base LLaVA-1.5-13B model on data generated with this approach are in \cref{tab:diff_gen_supp} (denoted as Baseline-II). We find that this method does well on some compositionality benchmarks (COLA and ConMe) but is not consistently better than the base LLaVA model especially on the control benchmarks. Qualitative examples of generated hard negatives from this approach and from the baseline swap objects/attributes approach along with SCRAMBLe's chain of thought approach are in \cref{tab:qual_eg_2}. We found that the feedback loop method could handle some more complex cases where a logical swap is not possible, but still the quality of generated captions is poorer that SCRAMBLe's chain of thought approach.

{  %
\renewcommand{\arraystretch}{1.5}  %
\begin{table*}[h]
    \centering
    \scalebox{0.75}{
    \begin{tabular}{@{}p{5cm}|p{5cm}|p{5cm}|p{5cm}@{}}
    \toprule
    \multicolumn{1}{c|}{\textbf{Positive Caption}} & \multicolumn{1}{c|}{\textbf{Baseline : Swap Obj/Att}} & \multicolumn{1}{c|}{\textbf{Baseline-II : Feedback Loop}} & \multicolumn{1}{c}{\textbf{SCRAMBLe : Chain of Thought}} \\
    \midrule
    A white horse pulling a cart down a street. & (Obj) A white cart pulling a horse down a street. & A cart is being pushed by a white horse up a street. & A white horse pushing a cart down a street. \\
    Close-up of bins of food that include broccoli and bread. & (Att) NA & Close-up of bins of food that exclude broccoli and include bread. & Wide shot of bins of food that include fruits and desserts \\
    A truck is pulling a horse trailer at a festival. & (Obj) A horse is pulling a truck trailer at a festival. & A horse is pulling a trailer instead of a truck at a festival. & A truck is being loaded with a horse trailer at a festival. \\
    Two women and a man posing for a photo on the dance floor. & (Obj) Two men and a woman posing for a photo on the dance floor. & Two women and a man posing for a photo off the dance floor. & One woman and two men posing for a photo on the dance floor. \\
    A yellow and blue fire hydrant that has fallen into the street with "caution" tape around it. & (Att) A blue and yellow fire hydrant that has fallen into the street with ``caution'' tape around it. & A yellow and blue fire hydrant stands upright that has not fallen into the street with "caution" tape around it & A yellow and blue fire hydrant near the sidewalk with 'caution' tape blocking off part of the street. \\
    A man with a white dress shirt black tie and a beard. & (Att) A man with a black dress shirt white tie and a beard. & A man with a black dress shirt and a white tie. & A man with a black dress shirt white tie and no beard. \\
    Three men in military suits are sitting on a bench. & (Obj) Three benches are sitting on military suits. & Three men in military suits are walking past a bench. & Three men in military suits are standing on a bench. \\
    A cake decorated to look like a female mouse. & (Obj) A mouse decorated to look like a female cake. & A cake decorated to look like a male mouse. & A cake decorated to look like a male mouse. \\
    Black and white photo of a man on the sidewalk pulling his luggage. & (Att) White and black photo of a man on the sidewalk pulling his luggage. & A man is pushing his luggage on the sidewalk in a black and white photo. & Color photo of a man on the sidewalk pushing his luggage. \\
    A yellow and white bed in a small room. & (Obj) A small bed in a yellow and white room. & A white bed in a large room with yellow walls. & A yellow and white bed in a large room. \\
    \bottomrule
    \end{tabular}
    }
    \caption{\textbf{Additional Qualitative Examples} of negative captions generated by different methods.}
    \label{tab:qual_eg_2}
\end{table*}
}  %

\section{Adversarial Refinement} \label{sec:app-adv_ref}

\begin{table}[t]
    \centering
    \scalebox{0.9}{
    \begin{tabular}{@{}lccc@{}}
    \toprule
    \textbf{LLaVA-1.5-13B   Perf} & \textbf{Winoground} & \textbf{EqBen} & \textbf{COLA} \\
    \midrule
    \textbf{w adv ref   (9.8k e.g.)} & 38.5 & 34.3 & 44.8 \\
    \textbf{w/o adv ref   (16.7k e.g.)} & 32.0 & 31.4 & 43.3 \\
    \bottomrule
    \end{tabular}
    }
    \caption{\textbf{Ablating Adversarial Refinement.} When our preference data has not been filtered using adversarial refinement, the performance of LLaVA-1.5-13B drops significantly on the compositionality benchmarks.}
    \label{tab:filter_ablation}
\end{table}

In \cref{sec:filtering} we described the adversarial refinement procedure to filter out examples for debiasing the preference tuning dataset using grammar and plausibility scores. 
The goal of this is that only based on plausibility or grammar scores of the captions (while disregarding the image) a model should not be able to correctly guess the positive caption over the negative (at any more than 50\% accuracy). We find that this debiasing is also effective for the preference tuning dataset, to avoid any model fitting to these biases. \cref{alg:ar} shows the adversarial refinement procedure.

\begin{algorithm}
    \small
      \caption{Adversarial Refinement}
      \label{alg:ar}  
      \begin{algorithmic}[1]  
       \Require 
          Grammar model $M_G$ and plausibility model $M_P$; Number of grids $K$;
          A set of candidates $\mathcal{D}=\left\{I_i, T^{\mathrm{p}}_i, T^{\mathrm{n}}_i\right\}_{i\in [N]}$, where $I_i$, $T^{\mathrm{p}}_i$, and $T^{\mathrm{n}}_i$ are $i$-th image, positive caption, and negative caption.
       \Ensure A subset $\bar{\mathcal{D}}\subset\mathcal{D}$
       \State Calculate the model score gap for each candidate $g^{(1)}_i=M_G(T^\mathrm{p}_i)-M_G(T^\mathrm{n}_i)$ and $g^{(2)}_i=M_P(T^\mathrm{p}_i)-M_P(T^\mathrm{n}_i)$
       \State Split the 2D space $[-1, 1]\times[-1, 1]$ to $K\times K$ equal-size grids.
       \State Place each candidate to a grid based on the score gaps $g^{(1)}_i$ and $g^{(2)}_i$.
       \State Initialize $\bar{\mathcal{D}}=\{\}$
        \For{each pair of grid $(G_j, G^{*}_j)$ symmetric about the original point $(0, 0)$}
            \If{$|G_j| > |G^*_j|$}
                \State Sample $|G^*_j|$ candidates from $G_j$ and put them to $\bar{\mathcal{D}}$. 
                \State Put candidates in $G^*_j$ to $\bar{\mathcal{D}}$. 
            \Else
                \State Sample $|G_j|$ candidates from $G^*_j$ and put them to $\bar{\mathcal{D}}$. 
                \State Put candidates in $G_j$ to $\bar{\mathcal{D}}$. 
            \EndIf
        \EndFor 
      \end{algorithmic}
\end{algorithm}  

In \cref{tab:filter_ablation} we show the performance of LLaVA-1.5-13B with and without adversarial refinement. We carry out this experiment by training the LLaVA-1.5-13B model on a smaller set of 16.7k examples from the COCO train set. After running adversarial refinement, we are left with 9.8k examples. Comparing performances of the two models, we see that tuning with the unfiltered data, causes performance on the compositionality benchmarks to drop significantly, indicating that adversarial refinement is crucial for retaining high quality examples for compositionality learning.

\section{Conversing with SCRAMBLe-Molmo : More Examples} \label{sec:app-chat_examples}

\begin{figure}[]
    \centering
    \includegraphics[width=\linewidth,trim=0cm 0cm 0cm 0cm,clip]{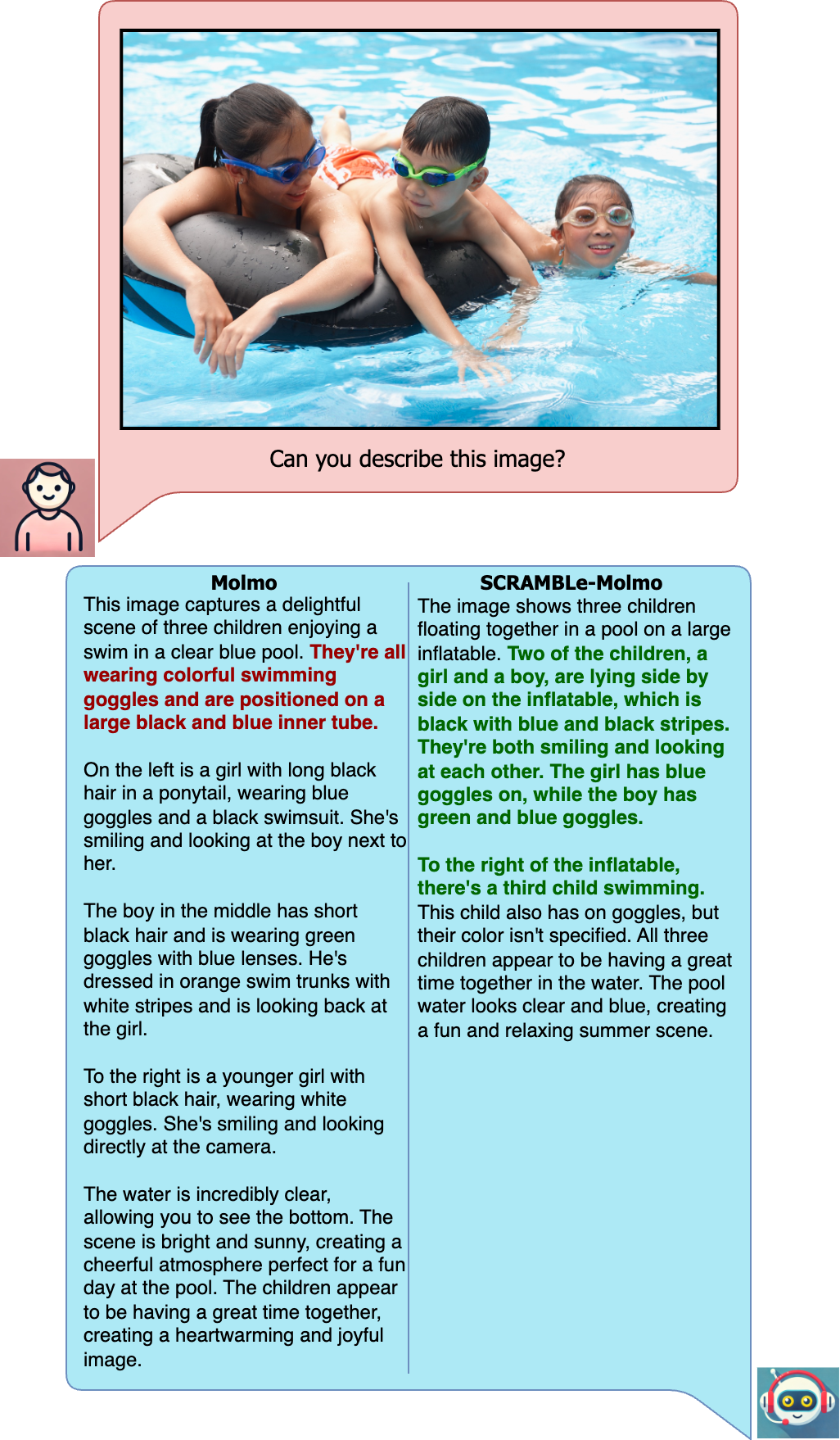}
    \caption{\textbf{Conversing with SCRAMBLe-Molmo (Winoground example).} In this example (which SCRAMBLe-Molmo gets right but Molmo does not), only SCRAMBLe-Molmo correctly identifies that only two of the kids are on the inflatable and the third one is not.}
    \label{fig:chat_eg4}
    \vspace{-2mm}
\end{figure}

\begin{figure}[]
    \centering
    \includegraphics[width=\linewidth,trim=0cm 0cm 0cm 0cm,clip]{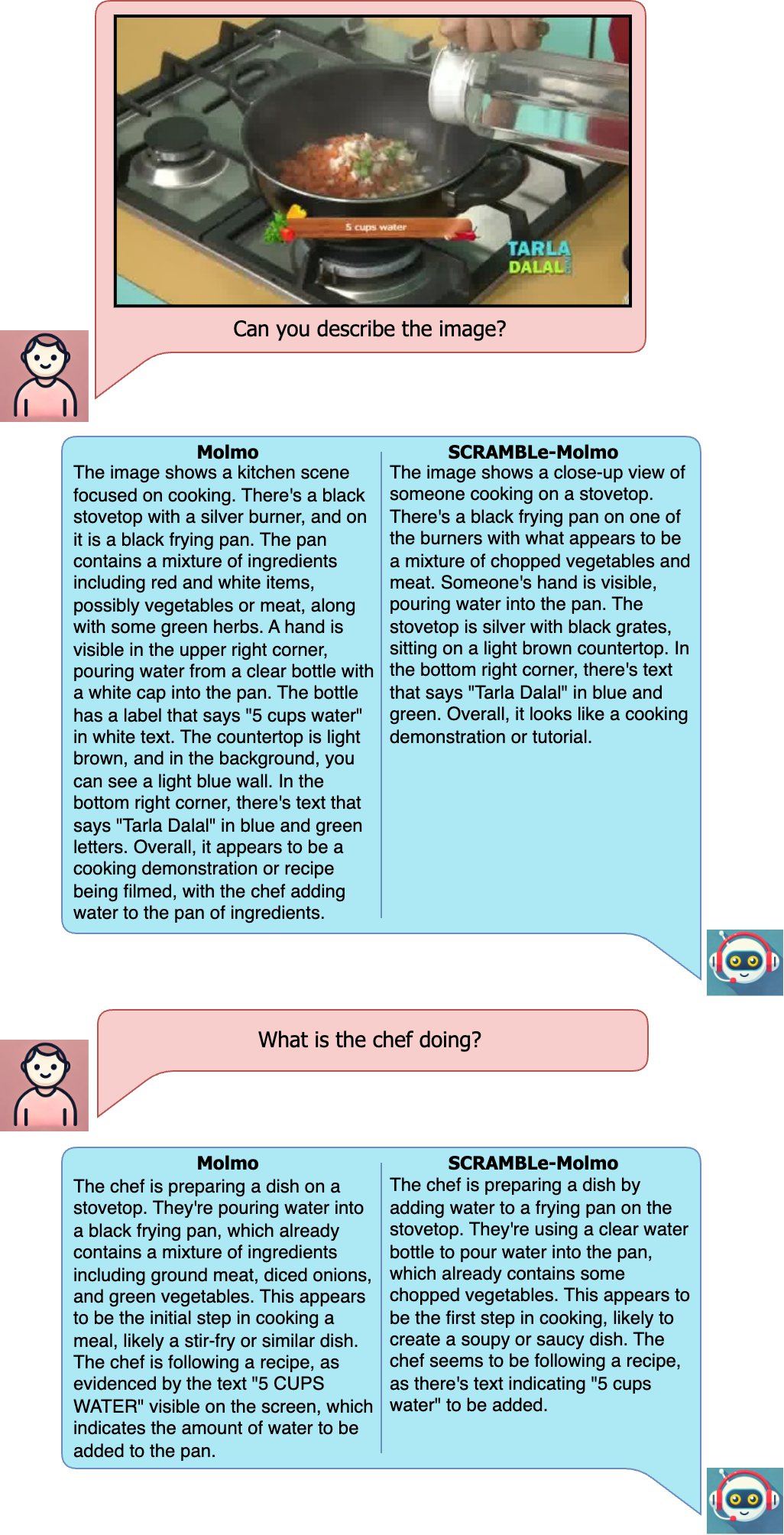}
    \caption{\textbf{Conversing with SCRAMBLe-Molmo (EQBen example).} The image is from an EQBen example that Molmo gets right but SCRAMBLe-Molmo does not. However, in conversation, SCRAMBLe-Molmo correctly identifies all elements of the scene and answers the question no worse than Molmo.}
    \label{fig:chat_eg3}
    \vspace{-2mm}
\end{figure}

\begin{figure*}[]
    \centering
    \includegraphics[width=\linewidth,trim=0cm 0cm 0cm 0cm,clip]{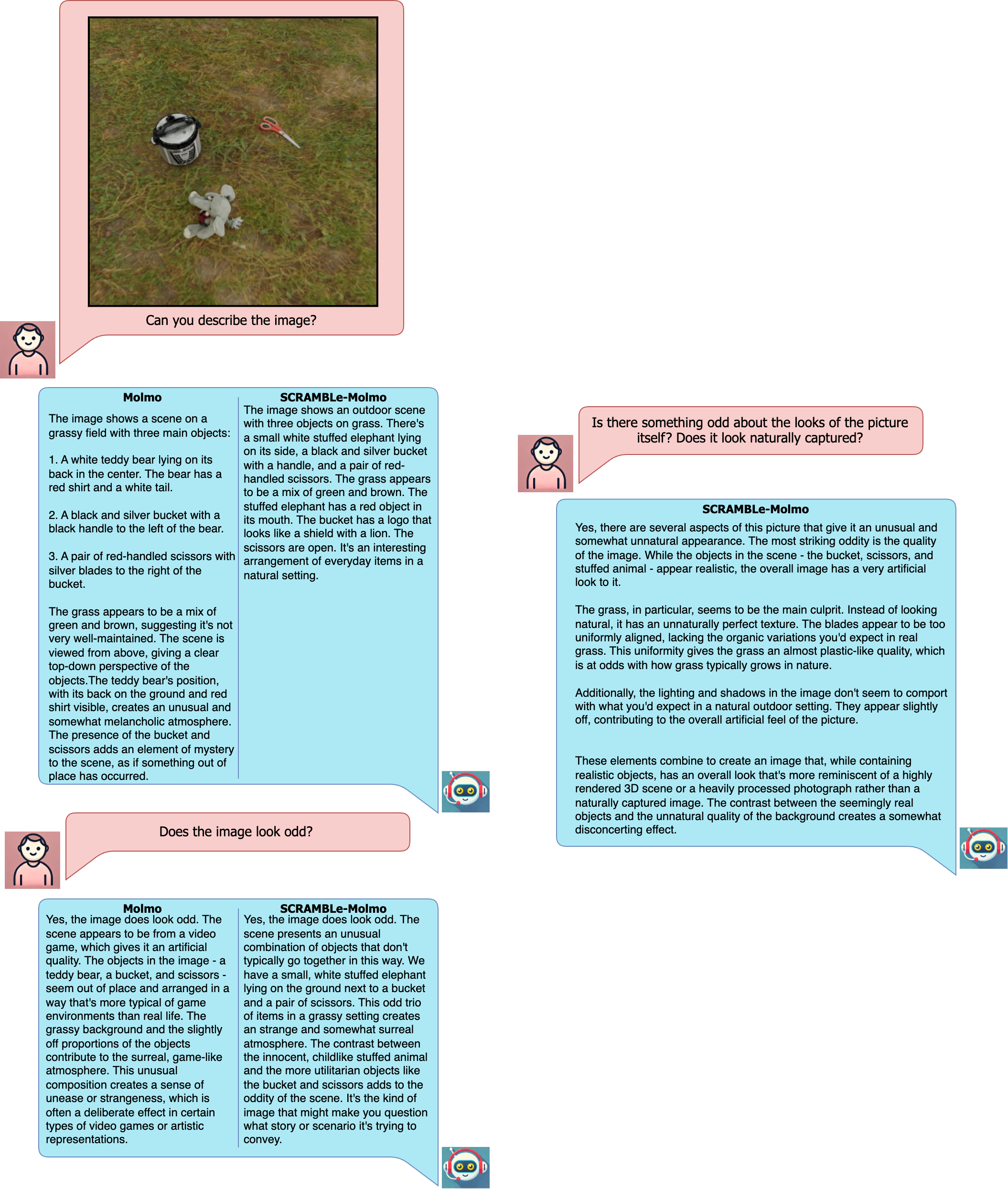}
    \caption{\textbf{Conversing with SCRAMBLe-Molmo (EQBen example).} This is another EQBen example that Molmo gets right but SCRAMBLe-Molmo does not. SCRAMBLe-Molmo does correctly identify the elements of the image. As a further question, when asked if there is something odd about the image, Molmo directly answers that the image looks artificially generated and game like, which SCRAMBLe-Molmo picks on the fact that the objects in the image do not typically appear together and it starts to read into the meaning of this. When probed further about the looks of the image however, it does mention that the image looks 3D rendered.}
    \label{fig:chat_eg5}
    \vspace{-2mm}
\end{figure*}

\begin{figure}[]
    \centering
    \includegraphics[width=\linewidth,trim=0cm 0cm 0cm 0cm,clip]{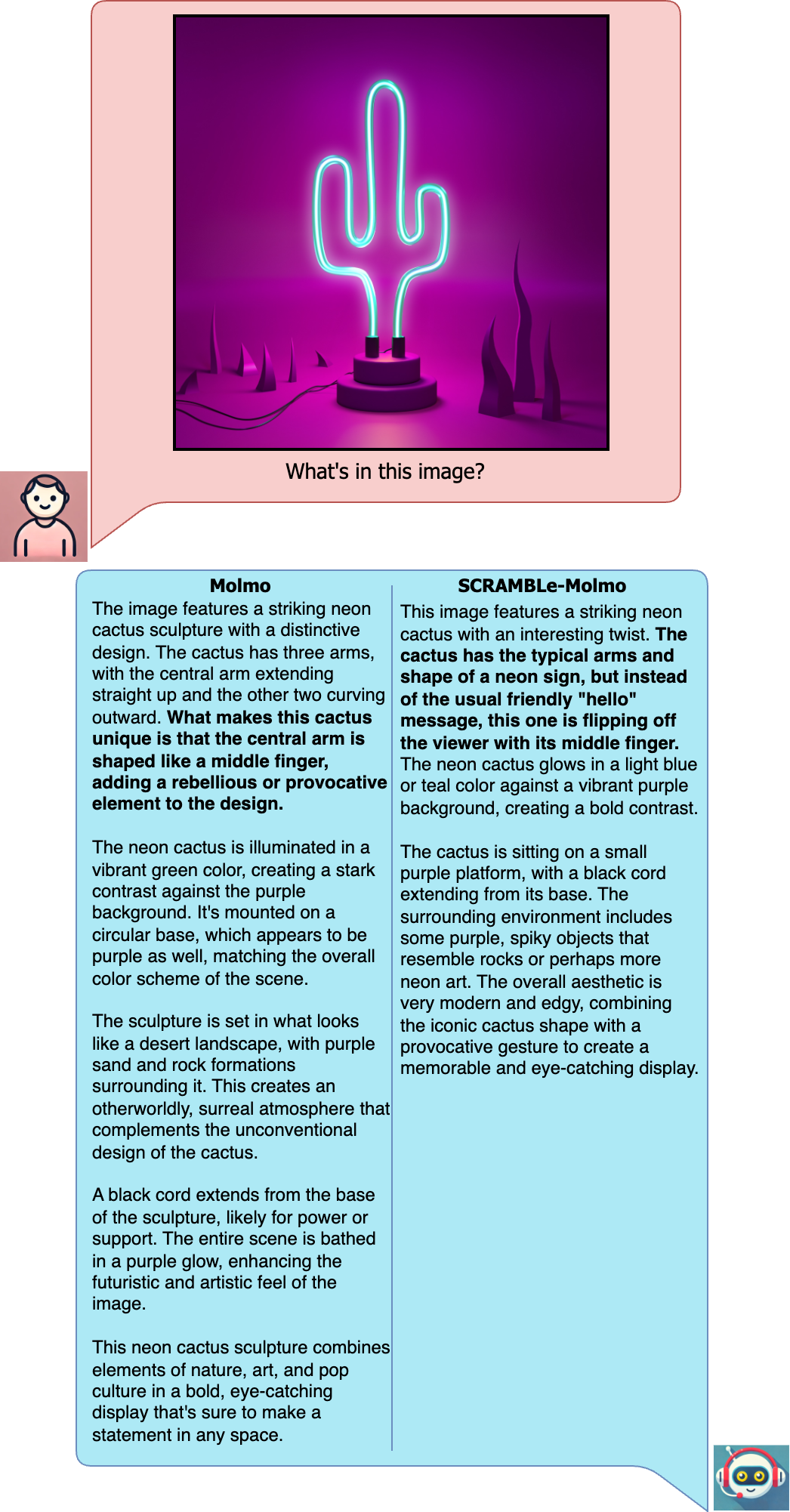}
    \caption{\textbf{Conversing with SCRAMBLe-Molmo (Winoground example).} Both Molmo and SCRAMBLe-Molmo seem to read into the symbolism that the middle arm of the cactus looks like a middle finger and gives the image a rebellious tone.}
    \label{fig:chat_eg6}
    \vspace{-2mm}
\end{figure}

\begin{figure}[]
    \centering
    \includegraphics[width=\linewidth,trim=0cm 0cm 0cm 0cm,clip]{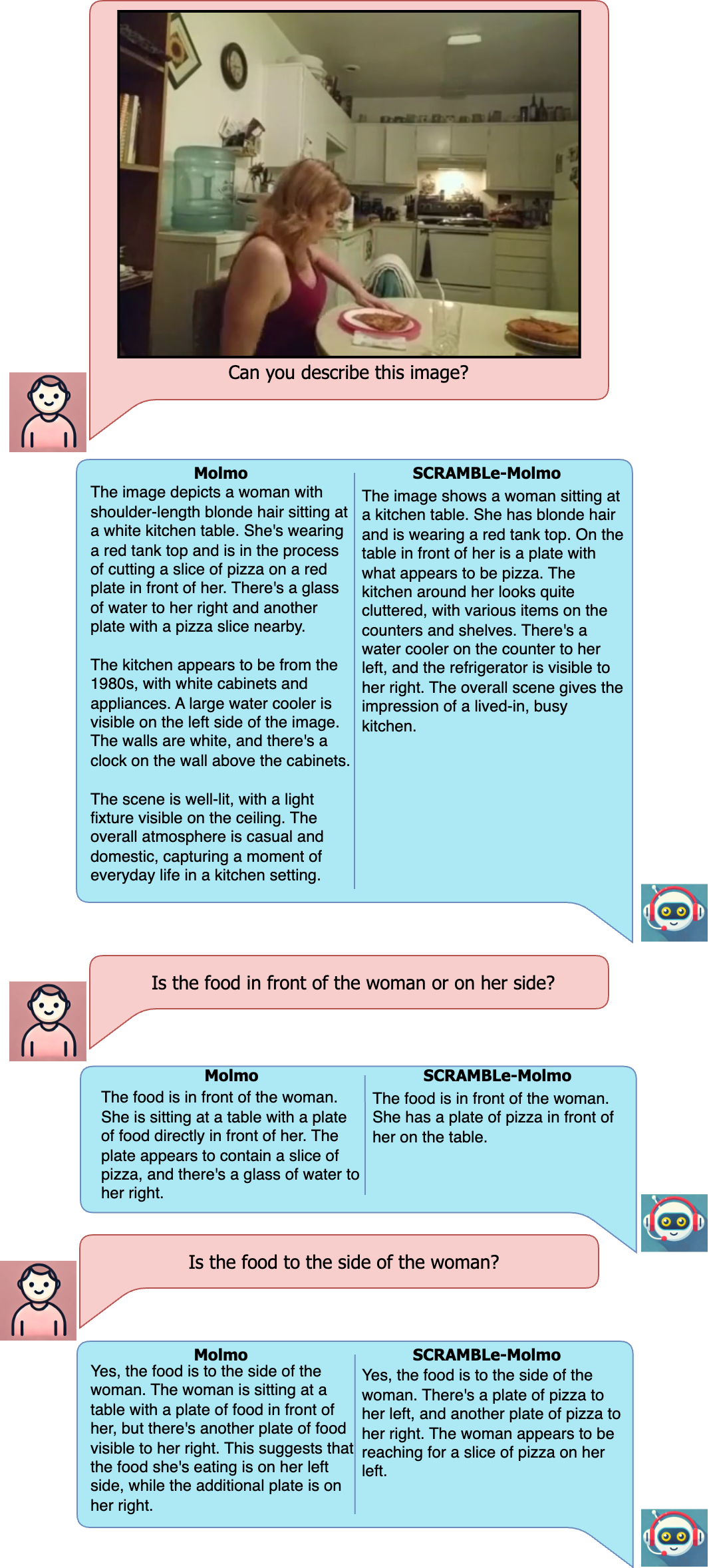}
    \caption{\textbf{Conversing with SCRAMBLe-Molmo (EQBen example).} This is another examples from EQBen that SCRAMBLe-Molmo gets wrong but Molmo gets right. On asking to describe the image, both models do a reasonable job. On the 2nd question too both models respond that the food is in front of the woman. Somewhat contradictorily though, on the last question, both models answer yes.}
    \label{fig:chat_eg7}
    \vspace{-2mm}
\end{figure}

More examples of conversive with Molmo and SCRAMBLe-Molmo are in \cref{fig:chat_eg3,fig:chat_eg4,fig:chat_eg5,fig:chat_eg6,fig:chat_eg7}. Please check the corresponding captions for more details.

\section{Implementation Details} \label{sec:app-impl}

All experiments in the paper were conducted on single Nvidia Ampere GPUs with a minimum 48G of VRAM (A100/A6000/A40/L40S/L40/RTX6000ada). We used the PyTorch framework~\cite{paszke2019pytorch} and our code for training MLLMs is based on Huggingface Transformers~\cite{wolf-etal-2020-transformers}, TRL~\cite{vonwerra2022trl} and POVID~\cite{zhou2024aligning}. We will upload our tuned models along with our synthetic data to Huggingface hub~\cite{huggingfacehub} along with the public release of our work. 

\subsection{Synthetic Data Generation.} As the LLM expert for synthetic caption generation, we used the Meta-Llama-3.1-70B-Instruct model. We ran inference at 4-bit quantization(nf4), with top-p sampling (p=0.9) and a temperature of 0.2. As our auxiliary grammar model we used textattack~\cite{morris2020textattack} and as the plausibility model we used Vera~\cite{liu2023vera}. These models were used both for filtering as well as feedback generation.

\subsection{Training}  \label{sec:app-training}

We used the direct preference optimization(DPO)~\cite{rafailov2024direct} objective for preference tuning, as described in \cref{sec:training}. As prescribed by Rafailov~\etal, we used a $\beta$ value of $0.1$. We trained each model with the AdamW optimizer, a base learning rate of \texttt{1e-5} and a cosine learning rate schedule with linear warmup for 3\% of the steps. 

\noindent\textbf{LLaVA.} For tuning the LLaVA-1.5-13B model, we trained for 2 epochs at a batch size of 8 (with no gradient accumulation). The rank of the low rank adapter (LoRA) was set to 32, with the $\alpha$ parameter set to 64 (this was selected from among candidate values $\{8, 16, 32, 64\}$ by validation performance over SugarCREPE-swap set). The base learning rate was \texttt{1e-5} and for the projector connecting the visual encoder to the language model we used a learning rate \texttt{2e-5}. The first stage of training took between 1.5-3 days to run (depending on the gpu used). For the second stage, we used a batch size of 1 and with 8 steps of gradient accumulation (for the effective batch size of 8; batch size was reduced to 1 because of the large memory footprint of some of the long LLaVA instruction tuning examples). We trained for 2 epochs with the same learning rates as stage 1. To prevent overfitting we used a label smoothing value of $0.1$ in the DPO loss in this stage. This stage took 5-10 hrs to run.

\noindent\textbf{Molmo.} We trained the Molmo-7B-D-0924 model for 2 epochs at a batch size of 2 and 4 gradient accumulation steps (for the same effective batch size of 8). The rank of LoRA was set to 16, with the $\alpha$ parameter set to 32. This took 1-1.5 days to train.

\noindent\textbf{Llama-3.2.} We trained the Llama-3.2-11B-Vision-Instruct model for 1 epoch at a batch size of 4 with 2 gradient accumulation steps (for the same effective batch size of 8). The rank of LoRA was set to 32, with the $\alpha$ parameter set to 64. In \cref{sec:results} we mentioned that we found this model to overfit to the full set of 57.8k synthetic examples. We hence trained this on a smaller set with 9.8k examples. This took around 12 hrs to train on an Nvidia A40 GPU.

\begin{table*}[h]
    \centering
    \scalebox{0.8}{
    \begin{tabular}{@{}lcccccc@{}}
    \toprule
    \textbf{Model} & \textbf{Winoground} & \textbf{EqBen} & \textbf{COLA} & \textbf{ConME} & \textbf{SEED-Bench} & \textbf{MM-Vet} \\
    \midrule
    \textbf{Llama-3.2-11B} & 31.5 & 43.6 & 37.1 & 71.3 & 13.79 & 57.0 $\pm$ 0.1 \\
    \textbf{+SCRAMBLe (57.8k eg)} & 34.3 & 43.6 & 33.8 & 70.1 & 27.79 & 35.0 $\pm$ 0.4 \\
    \textbf{+SCRAMBLe (9.8k eg)} & \textbf{35.3} & \textbf{44.3} & \textbf{40.0} & \textbf{74.6} & \textbf{42.74} & \textbf{60.3 $\pm$ 0.1} \\
    \bottomrule
    \end{tabular}
    }
    \caption{Llama-3.2-11B-Vision-Instruct on being tuned with the full set of 57k synthetic examples overfits to training data and leads to poorer performance across benchmarks.}
    \label{tab:llama3_overfit}
\end{table*}

In \cref{tab:llama3_overfit} we show the performance of the LLama-3 model on being trained with the full set of 58k synthetic examples. While the VQAScore evaluation on compositionality benchmarks still improves over the original Llama-3.2-11B-Vision-Instruct model, a benchmark like MM-Vet revealed some degenerate behaviors. Specifically, in this long answer generation task, the model trained on the full synthetic set often fell into loops of repeating a single phrase or a character, leading to drastic reduction in performance. We also note that the Llama-3.2 model performs poorly on SEED-Bench because it does not follow the format of the benchmark (responding to a multiple choice question with the letter corresponding to the correct answer) even when prompted to do so. This behavior improves a bit with our tuning, while even in this case, tuning with the smaller set is better.

\section{Synthetic Data Generating Conversations} \label{sec:synth_convo}

Here we show examples of the specific prompts/conversations with the LLM expert for the different methods of generation. In each of the conversations, the output of the LLM is colored in green. For the feedback loop, the output of the auxiliary feedback models is colored in blue.

\subsection{Baseline : Swap Objects.}
\begin{tcolorbox}[
    breakable,
    colback=gray!10,
    colframe=black,
    left=2pt,         %
    right=2pt,        %
    boxsep=2pt,        %
    sharp corners
]
\ttfamily
\fontsize{8}{11}\selectfont
Given an input sentence describing a scene, your task is to first locate two swappable noun phrases in the sentence, and then swap them to make a new sentence. The new sentence must meet the following three requirements:

\begin{enumerate}
    \item The new sentence must be describing a different scene from the input sentence.
    \item The new sentence must be fluent and grammatically correct.
    \item The new sentence must make logical sense.
\end{enumerate}

To complete the task, you should:
\begin{enumerate}
    \item Answer the question of whether generating such a new sentence is possible using Yes or No.
    \item Output the swappable noun phrases.
    \item Swap the selected noun phrases to generate a new sentence.
\end{enumerate}

Input: A woman cutting into a cake with a man standing behind her.

Is it possible to swap noun phrases in the input sentence to generate a new sentence that is different from the input sentence and makes logical sense? \textcolor[rgb]{0, 0.5, 0}{Yes.\\
Swappable noun phrases: a woman, a man\\
Output: A man cutting into a cake with a woman standing behind him.
}
\end{tcolorbox}
\label{fig:swap-obj-prompt-supp}

\subsection{Baseline : Swap Attributes.}
\begin{tcolorbox}[
    breakable,
    colback=gray!10,
    colframe=black,
    left=2pt,         %
    right=2pt,        %
    boxsep=2pt,        %
    sharp corners
]
\begin{flushleft}
\ttfamily
\fontsize{8}{11}\selectfont
\noindent
Given an input sentence describing a scene, your task is to first locate two swappable adjectives in the sentence describing different objects, and then swap them to make a new sentence.\par
\medskip
The new sentence must meet the following three requirements:\par
1. The new sentence must be describing a different scene from the input sentence.\par
2. The new sentence must be fluent and grammatically correct.\par
3. The new sentence must make logical sense.\par
\medskip
To complete the task, you should:\par
1. Answer the question of whether generating such a new sentence is possible using Yes or No.\par
2. Output the swappable adjectives.\par
3. Swap them to make a new sentence.\par
\medskip
Here are some examples:\par
\medskip
Input: A girl in a pink shirt holding a blue umbrella.\par
Is it possible to swap attributes in the input sentence to generate a new sentence that is different from the input sentence and makes logical sense? Yes\par
Swappable attributes: pink, blue\par
Output: A girl in a blue shirt holding a pink umbrella.\par

Input: A car and a truck are going through the intersection.\par
Is it possible to swap attributes in the input sentence to generate a new sentence that is different from the input sentence and makes logical sense? No\par
Swappable attributes: NA\par
Output: NA\par

Input: A cold drink on a hot day.\par
Is it possible to swap attributes in the input sentence to generate a new sentence that is different from the input sentence and makes logical sense? Yes\par
Swappable attributes: cold, hot\par
Output: A hot drink on a cold day.\par

Input: Four yellow airplanes flying side by side at an air show.\par
Is it possible to swap attributes in the input sentence to generate a new sentence that is different from the input sentence and makes logical sense? No\par
Swappable attributes: NA\par
Output: NA\par

Input: The dress on the left is long and the dress on the right is short.\par
Is it possible to swap attributes in the input sentence to generate a new sentence that is different from the input sentence and makes logical sense? Yes\par
Swappable attributes: long, short\par
Output: The dress on the left is short and the dress on the right is long.\par

Input: A big grey elephant standing in the jungle.\par
Is it possible to swap attributes in the input sentence to generate a new sentence that is different from the input sentence and makes logical sense? No\par
Swappable attributes: NA\par
Output: NA\par

Input: A large black dog is lying on a white comforter.\par
Is it possible to swap attributes in the input sentence to generate a new sentence that is different from the input sentence and makes logical sense? Yes\par
Swappable attributes: black, white\par
Output: A large white dog is lying on a black comforter.\par

Input: A red fire hydrant, a yellow balloon, and some rocks.\par
Is it possible to swap attributes in the input sentence to generate a new sentence that is different from the input sentence and makes logical sense? \color[rgb]{0, 0.5, 0}{%
Yes\par
Swappable attributes: red, yellow\par
Output: A yellow fire hydrant, a red balloon, and some rocks.
}%
\end{flushleft}
\end{tcolorbox}
\label{fig:swap-att-prompt-supp}

\subsection{SCRAMBLe : Chain of Thought}

\begin{tcolorbox}[
    breakable,
    colback=gray!10,
    colframe=black,
    left=2pt,         %
    right=2pt,        %
    boxsep=2pt,        %
    sharp corners
]
\begin{flushleft}
\ttfamily
\fontsize{8}{11}\selectfont
\noindent
Given an input caption describing a scene, your task is to rearrange words in it to make a new caption.\par
\medskip
The new caption must meet the following three requirements:\par
1. It must describe a scene with visual differences compared to the scene described by the input caption.\par
2. It must be fluent and grammatically correct.\par
3. It must make logical sense.\par
\medskip
Note that you can choose to abstain and output 'NA' if it is not possible to generate a negative caption for the given input.\par
\medskip
Here are some examples:\par
\medskip
Input Caption: a pink bird with a white beak\par
Reasoning:\par

1. Identify the key elements: \par
   - Color of the bird: pink\par
   - Color of the beak: white\par
   - Structure: "[color] bird with a [color] beak"\par

2. Recognize that the negative caption should describe a different visual image using the same words\par

3. Observe that the colors are the main distinguishing features\par

4. Swap the colors while maintaining the structure:\par
   - "pink" describing the bird becomes "white"\par
   - "white" describing the beak becomes "pink"\par

5. Keep the rest of the structure intact:\par
   - "a [color] bird with a [color] beak"\par

6. Apply the swapped colors to the structure:\par
   - "a white bird with a pink beak"\par
This transformation maintains the same words and grammatical structure but inverts the color assignments, creating a visually distinct image that serves as an effective negative caption.\par
Final Output Caption: a white bird with a pink beak\par
\medskip
Input Caption: a bottle is in water\par
Reasoning:\par

1. Identify the key elements:\par
   - Object 1: bottle\par
   - Object 2: water\par
   - Relationship: "is in"\par

2. Recognize that the negative caption should describe a different visual image using the same words\par

3. Observe that the relationship between the bottle and water is the distinguishing feature\par

4. Swap the positions of "bottle" and "water" in the sentence:\par
   - "Bottle" moves from the subject to the object position\par
   - "Water" moves from the object to the subject position\par

5. Adjust the articles as needed:\par
   - "a bottle" remains "a bottle" (no change needed)\par
   - "water" remains "water" (no article needed for this uncountable noun)\par

6. Keep the relationship "is in" intact\par

7. Reconstruct the sentence with the swapped elements:\par
   - "water is in a bottle"\par

This transformation maintains the same core words but inverts the spatial relationship between the bottle and water. It creates a contrasting scenario that effectively serves as a negative caption, describing a completely different visual image while using the same words.\par
Final Output Caption: water is in a bottle\par
\medskip
Input Caption: Manning a ship.\par
Reasoning:\par

1. Identify the key words: "manning" and "ship"\par

2. Recognize that "manning" is a verb form derived from "man"\par

3. Swap the roles of "man" and "ship":\par
   - "Man" becomes the object\par
   - "Ship" becomes the basis for the verb\par

4. Transform "ship" into a verb:\par
   - "Ship" becomes "shipping"\par

5. Change "man" to its object form:\par
   - "Man" becomes "a man"\par

6. Rearrange the words to form a grammatically correct sentence:\par
   - "Shipping a man"\par

This transformation maintains the same core words but completely changes the meaning and visual imagery, creating a contrasting scenario that serves as an effective negative caption.\par
Final Output Caption: Shipping a man.\par
\medskip
Input Caption: there are three bananas and two apples\par
Reasoning:\par

1. Identify the key elements:\par
   - Object 1: bananas\par
   - Object 2: apples\par
   - Quantity of bananas: three\par
   - Quantity of apples: two\par
   - Structure: "there are [number] [fruit] and [number] [fruit]"\par

2. Recognize that the negative caption should describe a different visual image using the same words\par

3. Observe that the quantities are the main distinguishing features\par

4. Swap the quantities while maintaining the fruits they describe:\par
   - "three" describing bananas becomes "two"\par
   - "two" describing apples becomes "three"\par

5. Keep the rest of the structure intact:\par
   - "there are" at the beginning\par
   - "and" connecting the two fruit phrases\par

6. Apply the swapped quantities to the structure:\par
   - "there are two bananas and three apples"\par

This transformation maintains the same words and grammatical structure but inverts the quantities assigned to each fruit. It creates a visually distinct image that serves as an effective negative caption, describing a different arrangement of fruits while using the exact same words as the positive caption.\par
Final Output Caption: there are two bananas and three apples\par
\medskip
Input Caption: There are more skiers than snowboarders.\par
Reasoning:\par

1. Identify the key elements:\par
   - Group 1: skiers\par
   - Group 2: snowboarders\par
   - Comparative structure: "more [group] than [group]"\par
   - Opening phrase: "There are"\par

2. Recognize that the negative caption should describe a different visual image using the same words\par

3. Observe that the comparative relationship between skiers and snowboarders is the distinguishing feature\par

4. Swap the positions of "skiers" and "snowboarders" in the sentence:\par
   - "skiers" moves from the subject of comparison to the object\par
   - "snowboarders" moves from the object of comparison to the subject\par

5. Keep the comparative structure "more ... than" intact\par

6. Maintain the opening phrase "There are"\par

7. Reconstruct the sentence with the swapped elements:\par
   - "There are more snowboarders than skiers"\par

This transformation maintains the same words and grammatical structure but inverts the comparative relationship between skiers and snowboarders. It creates a contrasting scenario that effectively serves as a negative caption, describing a completely different visual image (with snowboarders outnumbering skiers instead of vice versa) while using the exact same words as the positive caption.\par
Final Output Caption: There are more snowboarders than skiers.\par
\medskip
Input Caption: A white horse pulling a cart down a street.\par
Reasoning:\par

\color[rgb]{0, 0.5, 0}{%
1. Identify the key elements:\par
   - Object 1: horse\par
   - Color of the horse: white\par
   - Object 2: cart\par
   - Action: pulling\par
   - Location: down a street\par

2. Recognize that the negative caption should describe a different visual image using the same words\par

3. Observe that the action and the relationship between the horse and cart are the main distinguishing features\par

4. Swap the roles of the horse and cart:\par
   - The horse is no longer pulling the cart\par
   - Instead, the cart could be pulling the horse (although this is less common, we can use a different verb to make it more plausible)\par

5. Change the verb to make the new scenario more believable:\par
   - Instead of "pulling", use "being pulled by"\par

6. Keep the rest of the structure intact:\par
   - "A [color] [object] [verb] a [object] down a [location]"\par

7. Apply the changes to the structure:\par
   - "A white cart being pulled by a horse down a street" is not the best option, as it changes the color and type of the object being pulled. A better option would be to change the direction or the action.\par

Alternative solution:\par
   - Change the direction:\par
     - "A white horse pulling a cart up a street"\par
   - Change the action:\par
     - "A white horse pushing a cart down a street"\par

Final Output Caption: A white horse pushing a cart down a street.\par
\normalcolor}%
\end{flushleft}
\end{tcolorbox}
\label{fig:cot-prompt-supp}

\subsection{Baseline-II : Feedback Loop} \label{sec:feedback_convo}

\begin{tcolorbox}[
    breakable,
    colback=gray!10,
    colframe=black,
    left=2pt,         %
    right=2pt,        %
    boxsep=2pt,        %
    sharp corners
]
\begin{flushleft}
\ttfamily
\fontsize{8}{11}\selectfont
\noindent
Given an input caption describing a scene, your task is to rearrange words in it to make a new caption. The new caption must meet the following three requirements:\par
\medskip
1. It must describe a scene with visual differences compared to the scene described by the input caption.\par
2. It must be fluent and grammatically correct.\par
3. It must make logical sense.\par
\medskip
Note that you can choose to abstain and output 'NA' if it is not possible to generate a negative caption for the given input.\par
\medskip
To help with your task, I will rate your output based on grammar (0-1), plausibility (0-1), and whether there are visual differences between the original caption and your output (Yes/No).\par
\medskip
In your output, please follow the format\par
\medskip
Final Output Caption: <caption>.\par
\medskip
Input: A white horse pulling a cart down a street.\par
\textcolor[rgb]{0, 0.5, 0}{Final Output Caption: A cart is being pulled down the street by a white horse.}\par
\medskip
FEEDBACK:\par
Your grammar score is \textcolor[rgb]{0, 0, 0.5}{0.99}.\par
Your plausibility score is \textcolor[rgb]{0, 0, 0.5}{0.56}.\par
Is the output caption visually different from the original caption? : \textcolor[rgb]{0, 0, 0.5}{No}\par
Your output caption has extra words (lemmatized): \textcolor[rgb]{0, 0, 0.5}{'the', 'by', 'be'}.\par
Can you please try again?\par
\medskip
\textcolor[rgb]{0, 0.5, 0}{Final Output Caption: A cart is pulled by a horse down the street.}\par
\medskip
FEEDBACK:\par
Your grammar score degraded to \textcolor[rgb]{0, 0, 0.5}{0.99}.\par
Your plausibility score improved to \textcolor[rgb]{0, 0, 0.5}{0.88}.\par
Is the output caption visually different from the original caption? : \textcolor[rgb]{0, 0, 0.5}{No}\par
Your output caption has extra words (lemmatized): \textcolor[rgb]{0, 0, 0.5}{'the', 'by', 'be'}.\par
Your output caption has missing words (lemmatized): \textcolor[rgb]{0, 0, 0.5}{'white'}.\par
Can you please try again?\par
\medskip
\textcolor[rgb]{0, 0.5, 0}{Final Output Caption: A horse is not pulling a cart down a street.}\par
\medskip
FEEDBACK:\par
Your grammar score degraded to \textcolor[rgb]{0, 0, 0.5}{0.98}.\par
Your plausibility score degraded to \textcolor[rgb]{0, 0, 0.5}{0.24}.\par
Is the output caption visually different from the original caption? : \textcolor[rgb]{0, 0, 0.5}{Yes}\par
Your output caption has extra words (lemmatized): \textcolor[rgb]{0, 0, 0.5}{'not', 'be'}.\par
Your output caption has missing words (lemmatized): \textcolor[rgb]{0, 0, 0.5}{'white'}.\par
Can you please try again?\par
\medskip
\textcolor[rgb]{0, 0.5, 0}{Final Output Caption: A white horse is standing next to a cart on a street.}\par
\medskip
FEEDBACK:\par
Your grammar score improved to \textcolor[rgb]{0, 0, 0.5}{0.99}.\par
Your plausibility score improved to \textcolor[rgb]{0, 0, 0.5}{0.34}.\par
Is the output caption visually different from the original caption? : \textcolor[rgb]{0, 0, 0.5}{Yes}\par
Your output caption has extra words (lemmatized): \textcolor[rgb]{0, 0, 0.5}{'next', 'stand', 'to', 'on', 'be'}.\par
Your output caption has missing words (lemmatized): \textcolor[rgb]{0, 0, 0.5}{'down', 'pull'}.\par
Can you please try again?\par
\medskip
\textcolor[rgb]{0, 0.5, 0}{Final Output Caption: A cart is being pushed by a white horse up a street.}\par
\medskip
FEEDBACK:\par
Your grammar score improved to \textcolor[rgb]{0, 0, 0.5}{0.99}.\par
Your plausibility score improved to \textcolor[rgb]{0, 0, 0.5}{0.45}.\par
Is the output caption visually different from the original caption? : \textcolor[rgb]{0, 0, 0.5}{Yes}\par
Your output caption has extra words (lemmatized): \textcolor[rgb]{0, 0, 0.5}{'up', 'push', 'by', 'be'}.\par
Your output caption has missing words (lemmatized): \textcolor[rgb]{0, 0, 0.5}{'down', 'pull'}.\par
\end{flushleft}
\end{tcolorbox}
\label{fig:feedback-convo-supp}

\break

\end{document}